\documentclass[10pt, a4paper]{article}

\usepackage[final]{lrec2026} % this is the new style
\usepackage{booktabs}
\usepackage{xspace}
\usepackage{multirow}
\usepackage{array}
\usepackage[shortlabels]{enumitem}

\newcommand{\tone}{$t_1$\xspace}
\newcommand{\ttwo}{$t_2$\xspace}
\newcommand{\ignore}[1]{}

\usepackage{layout}

\newcommand{\camerareadytext}[1]{\ignore{#1}}
\title{What Are LLMs Doing to Scientific Communication?\\Measuring Changes in Writing Practices and Reading Experience}

\name{Filip Miletić\footnotemark[1]{*}\hspace{1cm}Neele Falk\footnotemark[1]{*}} 

\address{Institute for Natural Language Processing,
         University of Stuttgart, Germany \\
         \{filip.miletic, neele.falk\}@ims.uni-stuttgart.de \\}
\abstract{
Has the style of scientific communication changed due to the growing use of large language models in the writing process? 
We address this question in the domain of Natural Language Processing by leveraging two data resources we create: a naturalistic corpus of over 37,000 papers from the ACL Anthology (2020--2024);
and a synthetic dataset of 3,000 human-written passages and their LLM-generated improvements.
We first implement a series of diachronic lexical analyses, showing that both word frequency and usage contexts have changed significantly over time, indicating semantic specialization in some cases and generalization in others.
Broadening our perspective, we then model a range of more complex stylistic features and find that LLM-modified texts more frequently contain certain syntactic constructions, more complex and longer words and a lower lexical diversity.
Finally, we connect these changes in writing practices to subjective reading experience through a pilot annotation study with 20 domain experts. 
They overall rate LLM-improved texts as more understandable and exciting, but also express negative qualitative attitudes towards LLMs, highlighting the strongly subjective effect of AI-assisted writing on reading experience.
 \\ \newline \Keywords{AI-assisted writing, scientific communication, language change} }

\begin{document}

\maketitleabstract

\renewcommand{\thefootnote}{\fnsymbol{footnote}}
\footnotetext[1]{Equal contribution.}
\renewcommand{\thefootnote}{\arabic{footnote}}

\section{Introduction}
%\red{For shortening text that we want to keep for later: you can make text vansih with command camerareadytext}

Large language models (LLMs) are increasingly used to assist human writing, including in high-stakes domains such as scientific communication.
The rapid and pervasive nature of these changes raises the question of the ways in which they may be altering prevalent writing practices (e.g., lexical and stylistic choices),
and of the subsequent effect of those practices on reading experience (e.g., perceived clarity and trustworthiness of texts).
%The consequential nature of these processes is supported by the hypothesis that widespread LLM use may lead to the development of new language varieties \citep{grieve2025sociolinguistic} as well as an emergent body of empirical accounts (see Section~\ref{sec:related-work}).

While evidence of these changes is beginning to emerge, prior work is limited in two major respects.
First, several recent studies examine the distinctive linguistic properties of LLM-generated scientific text \cite{Ma23, MuozOrtiz2024, zanotto24, zamaraeva-etal-2025-comparing}, but they generally compare texts which are written by humans vs.\ entirely model-generated.
This clear-cut distinction oversimplifies finer-grained practices:
LLMs are typically used to improve human-written text rather than generate entire passages \cite{Koller2024,Kobak2025};
moreover, real-life documents tend to alternate between human-only and LLM-improved writing rather than contain a uniform amount of generated text \cite{coauthor,richburg-etal-2024-automatic}.
%such model-improved text appears in only some parts of a document rather than being uniformly distributed throughout \red{(reference)}.
The second major shortcoming is the focus on identifying distinctive properties of generated text without systematically measuring their effect on human readers.
Even where included, such measurements are limited to broad patterns such as the ability to distinguish human vs.\ LLM writing \cite{Gao2023,Ma23}.
As a result, the link between objective differences in writing style and more subjective but vital dimensions of reading experience %-- such as the perceived clarity and trustworthiness of a text -- 
remains to be established.

This paper aims to provide a more realistic and comprehensive assessment of LLM use in scientific communication.
We design our study so as to capture the collaborative nature of human--LLM writing and the uneven spread of such interventions within a document, as well as to explicitly connect the distinctive features of these writing practices with subjective reading experience.
%The present paper aims to provide a more realistic assessment of generative models in scientific communication -- capturing the collaborative nature of human-LLM writing and the uneven spread of those interventions within a document -- as well as to explicitly connect our insights into stylistic differences with subjective reading experience.
We conduct our analysis on NLP papers from the ACL Anthology, and define two periods of around two years each, respectively preceding and following the release of ChatGPT in November 2022.
We view the two periods as reflecting community-level writing practices which do not include any vs.\ may include some writing by public-facing LLMs.
Complementing this naturalistic scenario, we simulate the real-life use of LLMs in a more controlled synthetic setting: we sample 3,000 extracts from pre-ChatGPT papers and generate model-improved versions of those.
We pose the following research questions:

\noindent
\begin{tabular}{@{}p{0.11\linewidth}@{}p{0.89\linewidth}}
    \textbf{RQ1} & In what way did core linguistic choices change between these two periods?\\
    \textbf{RQ2} & To what extent are more complex stylistic properties specific to each of the two periods?\\
    \textbf{RQ3} & Do these differences in writing practices give rise to different reading experiences?\\
\end{tabular}

We first assess differences in linguistic choices by deploying a series of diachronic lexical analyses: statistical corpus measures to identify emergent terms; type-level word embeddings to characterize broad changes in their semantic properties; and token-level word embeddings to automatically retrieve their time-specific uses.
Broadening our focus, we then investigate how different linguistic features — such as text length, sentiment, grammatical and lexical variability, and readability — contribute to explaining variation between human and LLM-assisted writings through regression analysis.
Finally, we conduct an annotation study contrasting human-written texts with their LLM-produced improvements, and ask 20 domain experts for ratings of reading experience in terms of clarity, authenticity, trustworthiness, and excitement.

We provide the following contributions.
\textbf{(1)}~We show that post-ChatGPT papers are distinguished by more complex lexical choices (e.g., \textit{enhance} rather than \textit{improve}) and further stylistic properties (e.g., lower lexical diversity).
By comparing naturalistic data from the ACL Anthology and synthetic data from text generation experiments, we confirm that these writing practices can be attributed to LLM use.
\textbf{(2)}~We further find that these stylistic changes are linked to differences in subjective reading experience, with LLM-improved texts perceived as clearer and more exciting.
\textbf{(3)}~We release an updated version of the ACL-OCL corpus \citep{rohatgi-etal-2023-acl}, containing PDF-extracted text of 99.3k papers from the ACL Anthology. We also provide a one-line script to ingest future papers. %and a trained topic model that can be run over future updates.
\textbf{(4)}~We release 3,000 pairs of human-written texts and their LLM-produced improvements, and annotations of human reading experience for 200 pairs.%
\footnote{Data and code are available at \url{https://github.com/FilipMiletic/ScientificCommunication}}

\section{Related Work} \label{sec:related-work}
\paragraph{Detection of AI-generated content.}
\camerareadytext{
In general, the demand for good tools for detecting AI-generated content and the interest in researching different methods has grown significantly since the success of ChatGPT. Because the models are constantly being updated, it is difficult to find methods that generalize well. For the development of models, related works have released large datasets that contain both human-written and LLM-generated texts, e.g. OpenLLMText \cite{chen-etal-2023-token}, MAGE \cite{li-etal-2024-mage} or HC3 \cite{HC3}), as well as benchmarks that test how well different models are recognized \cite{dugan-etal-2024-raid}, some of which also include languages other than English (MULTITuDE \cite{macko-etal-2023-multitude}, M4 \cite{wang-etal-2024-m4}).
Methods for detecting AI-generated content include techniques that are directly integrated into the models, such as watermarking \cite{watermarking}, fine-tuning transformer-based classifiers \cite{Guggilla2025AIGT}, or statistical methods that make use of model-related features, such as logits or word probabilities (see \citet{wu-etal-2025-survey}, for a comprehensive overview of existing datasets and methods).
Of particular interest for this work are approaches that make predictions based on linguistic features. For example,  \citet{hamedWu} found that the vocabulary of ChatGPT generated articles is much less diverse than in human writings.}

%Since the popularity of LLMs, such as ChatGPT, has increased over the last years, research on how to automatically detect AI-generated content has grown significantly.
Paralleling the rise in popularity of LLMs, recent years have seen growing interest in automatic detection of AI-generated content.
This includes the release of datasets and benchmarks to train detection tools and evaluate different methods 
\citep[e.g.,][]{chen-etal-2023-token, li-etal-2024-mage, HC3, dugan-etal-2024-raid, macko-etal-2023-multitude, wang-etal-2024-m4}. 
%e.g., OpenLLMText \cite{chen-etal-2023-token}, MAGE \cite{li-etal-2024-mage}, HC3 \cite{HC3}, RAID \cite{dugan-etal-2024-raid}, MULTITuDE \cite{macko-etal-2023-multitude}, and M4 \cite{wang-etal-2024-m4}. 
Techniques for detecting AI-generated text include watermarking \cite{watermarking}, fine-tuning transformer-based classifiers \cite{Guggilla2025AIGT}, using model-related features \citep{wu-etal-2025-survey} or linguistic features \citep{hamedWu}.\footnote{We refer to the survey by \citet{wu-etal-2025-survey} for a comprehensive overview on detecting AI-generated content.}
While good results are generally achieved for AI-generated content, the detection of human--AI coauthored text remains a major challenge and requires adaptation of existing models \cite{richburg-etal-2024-automatic, su-etal-2025-haco}. Early work includes the CoAuthor dataset \cite{coauthor}, which includes essays augmented with GPT-3 suggestions, while more recent datasets focus on different variations of human–AI co-authored texts (e.g., human-written then machine-polished) \cite{coauthorbenchmark}.

%Opposed to previous findings, they find that AI-modified text exhibits a greater lexical diversity compared to humans only.

\paragraph{Stylistic differences between human-written and AI-generated or AI-modified text.}

% integrate this somewhere
\camerareadytext{\citet{Pan2025} explore authorship attribution of human-AI co-authored text and in that context analyze the role of stylistic features. Opposed to previous findings, they find that AI-modified text exhibits a greater lexical diversity.} %compared to humans only.
Several works more directly explore the difference in stylistic features of human and AI-generated text. Domains that are mostly covered are news articles \cite{MuozOrtiz2024,zamaraeva-etal-2025-comparing}, essays \cite{Akinwande_2024} and abstracts of scientific articles \cite{Ma23}.
Existing works examine features from all possible categories, such as the frequency of certain syntactic constructions, n-grams, hedging, lexical complexity, rhetorical properties and sentiment.
Frequent linguistic peculiarities in AI-generated content include, e.g., lower lexical variation \cite{zanotto24,Akinwande_2024,YildizDurak2025}, more positive sentiment \cite{MuozOrtiz2024,zamaraeva-etal-2025-comparing}, fewer compounds \cite{zamaraeva-etal-2025-comparing}, and the excessive use of certain verbs and modifiers such as \textit{delve}, \textit{crucial}, or \textit{intricate} 
\cite{Gray2024ChatGPTE,Kobak2025,Reinhart2025}.

Several works use these features to predict whether a text was human or AI-generated and to identify the strongest predictor \cite{Ma23,DESAIRE23,Akinwande_2024}. %These classifiers frequently yield a high performance. % human
Some works also investigate human perception of LLM-generated texts, e.g. \citet{Gao2023} and \citet{Hakam2024-ql} find that human annotators struggle to distinguish between human and LLM-generated scientific texts.  \citet{russell-etal-2025-people} show that annotators with frequent LLM-writing experience better detect generated news. In \citet{Doru2025}, participants classified scientific texts and rated their fluency, quality, and coherence. \citet{Lin25} find that researchers use LLMs mainly to improve clarity and conciseness, leading to a more homogeneous writing style since ChatGPT’s release.

\camerareadytext{Existing work has focused considerably on the stylistic and lexical differences between LLM-generated and human texts, including in the scientific domain. However,} 
\camerareadytext{The cited studies often analyze completely generated texts (text continuation or paraphrasing) and rarely compare hybrid (human-AI co-authored) texts with purely human texts. For this reason, we compare articles that appeared after ChatGPT with those that appeared shortly before its release. Even if only a fraction of the post-ChatGPT articles are LLM-modified, certain linguistic patterns should occur (albeit weaker than in the synthetic case).
Further, LLM writing style may be unconsciously adopted by researchers who frequently interact with LLM-modified language.}
Most prior studies focus on fully generated texts and rarely compare human–AI co-authored vs.\ human-only writing. For this reason, we compare articles published after ChatGPT’s release with those from shortly before, expecting weaker but detectable linguistic shifts even if only a fraction were LLM-modified. Further, frequent exposure to LLM-generated language may lead researchers to unconsciously adopt its style.
Unlike most prior studies, we analyze full papers rather than abstracts, since LLM use likely occurs across all sections. While LLM-related vocabulary and linguistic features have been studied, they are rarely examined together, and existing research often focuses on surface-level trends. Lexical choices, in particular, remain underexplored beyond frequency-based analyses, despite well-established methods for modeling semantic change \citep{tahmasebi2021survey, Schlechtweg2023measurement}.
Finally, the subjective perception of LLM-generated content in scientific texts has hardly been studied, which is why we complement our data-driven analysis with a pilot study of reading experience with 20 domain experts.

%it is possible that the writing style of LLMs is unconsciously adopted, as researchers frequently interact with LLM-modified or generated language. 
\camerareadytext{Unlike most prior work on the scientific domain, we do not restrict ourselves to abstracts but rather  examine complete papers, as LLM use is to be expected in all paper sections. %; we focus on the NLP community.
With regard to linguistic features, LLM-frequent vocabulary and certain linguistic features have been studied, but rarely both together. Furthermore, many current findings rely on surface-level trends. In particular, lexical choices are yet to be analyzed beyond rates of use in terms of semantic change, %the semantic change in word usage before and after LLM influence has not yet been investigated 
despite the wide availability of methods to model diachronic phenomena \citep{tahmasebi2021survey, Schlechtweg2023measurement}.}

\section{Data}
We now present our two English-language data resources: a naturalistic corpus of NLP papers from the ACL Anthology (henceforth \textit{original} dataset); and a synthetic dataset of human-written passages and their LLM-generated improvements (henceforth \textit{LLM} dataset).

\subsection{ACL Anthology Corpus}
\label{sec:acl-antho-corpus}
Since the focus of our work is on scientific communication in the NLP community, we analyze the papers from the ACL Anthology,%
\footnote{\url{https://aclanthology.org}}
the open-access publication repository of the Association for Computational Linguistics (ACL).
% \camerareadytext{The Anthology was founded in 2002 and hosts the proceedings of ACL-sponsored events as well as those organized by its sister organizations such as the ELRA. %\citep{gildea-etal-2018-acl}.
% With over 113,000 papers available as of this writing, the Anthology exhaustively documents not only the scientific findings of the NLP community, but also the evolution of its writing practices.}
As our starting point, we use the ACL-OCL corpus \citep{rohatgi-etal-2023-acl}
containing ca.\ 73,000 papers. %published between 1952 and 2022.
They were obtained by crawling the Anthology website for PDF files and then extracting the full text using GROBID.%
\footnote{\url{https://github.com/kermitt2/grobid}}
% The corpus was further enriched with paper metadata, %(from the Anthology website), 
% citation counts, %(from SemanticScholar), 
% and topic labels. %obtained by training a transformer-based classifier.
% However, we note several recurrent issues limiting the utility of the corpus for our study.
% We therefore implement additional data processing steps to produce CACL-OCL, the clean and updated version of the corpus.%
% \footnote{To be publicly released upon acceptance.}
%
%\subsubsection{Corpus expansion and cleaning}
% We update the corpus to extend its temporal coverage to the present day and resolve other recurrent issues noted in the original version.
%
%\paragraph{Extending data coverage}

The content in the original corpus ends in September 2022 and to our knowledge has not been updated since.
We therefore implement an update to bring its temporal span closer to the present day.
We also note two other recurrent problems.
Some papers from the original time span are available in the Anthology but not included in the corpus, possibly due to coverage issues during crawling.
Other papers are included in the corpus, but are associated with metadata without full text content due to file issues (e.g., failed extraction with GROBID or PDF missing from the Anthology at the time of the crawl, especially for very early conferences).

In our update,
we do not crawl the Anthology website but instead use its BibTeX export as the most comprehensive structured record of available papers.
We rely on BibTeX information to identify the missing papers based on citation keys, extract their metadata, and reconstruct their URLs.
We download the corresponding PDF files and then use GROBID  to extract paper text.
This process also recovers the textual content for a subset of papers lacking it in the original corpus; we remove any remaining papers without textual content.
We accompany the updated corpus with code (run as a one-line command) which checks the locally available papers against those in the Anthology and passes any missing papers through the full update pipeline. 

The updated corpus contains 99.2k papers published until the end of 2024.
For the purposes of our study, we define a subcorpus structured into two time periods around a critical point in time regarding LLM use: the release of ChatGPT in November 2022.
The first time period (\tone) contains papers published from 2020 to 2022.
Its last major conference event is EMNLP~2022, which had a camera-ready deadline in October of that year.
The second time period (\ttwo) contains papers published in the second half of 2023 and in 2024.
It begins with ACL 2023, whose camera-ready deadline was in May of that year, i.e., six months after the release of ChatGPT.
The gap between the two periods ensures a clear distinction between them while limiting the effect of changes in topic over time.
We only retain papers from events that took place in both time periods.  %, and we exclude front-matter content such as volume introductions.
% The only major venue excluded in the six-month gap is EACL~2023.
% ChatGPT became available after the initial submissions but before the camera-ready deadline, so we view this conference as belonging to a transition period with unclear effects on writing. 

While we assume that these sampling constraints ensure good comparability of \tone and \ttwo, we also inspect it more closely by running a topic analysis.
We find some topical shifts in line with the evolution of the field (e.g., a stronger focus on individual levels of linguistic structure in \tone, and prominence of more recent machine learning methods in \ttwo),
but the analysis overall confirms broad topical comparability of the two time periods.
Detailed results are reported in Appendix~\ref{app:topic}.

We preprocess PDF-extracted text using spaCy\footnote{\url{https://spacy.io}} (model \texttt{en\_core\_web\_sm}). We segment the text into sentences, which are then tokenized, lemmatized, and POS-tagged. We run a subset of analyses on paragraph-level, which we operationally define as non-overlapping windows of five sentences.
Final corpus structure is shown in Table~\ref{tab:paper-distribution}.

\begin{table}[t]
    \centering
    \resizebox{\linewidth}{!}{%
    \begin{tabular}{lrrrr}
        \toprule
        & \multicolumn{1}{c}{\textbf{Time period}} & \textbf{Papers} & \textbf{Paras} & \textbf{Tokens}\\
        \midrule
        \tone & Jan 2020 -- Dec 2022 & 20,259 & 801,016 & 100.2\,m \\
        \ttwo & Jul 2023 -- Dec 2024 & 17,501 & 831,077 & 103.6\,m \\
        \midrule
        & Total & 37,760 & 1,632,093 & 203.8\,m\\
        \bottomrule
    \end{tabular}}
    \caption{Distribution of papers across time periods} %(after excluding non-English content and defective PDF conversions)
    \label{tab:paper-distribution}
\end{table}

\subsection{LLM-Assisted Paraphrases}

The original sub-corpus described in Section \ref{sec:acl-antho-corpus} describes the  realistic scenario in which \ttwo contains a hybrid form of human and LLM co-authored text.
To compare whether the patterns that emerge from the analysis on this are similar to those in texts explicitly modified by LLMs, we replicate this scenario in a controlled setup and construct a dataset with texts from \tone paired with GPT-improved paraphrases, thus offering a clear gold label (human vs. LLM-modified).

We select a random sample of 3,000 publications from \tone from 2022 and, for each paper in this, a random paragraph with a minimum length of 100 tokens. The selected paragraph is chosen from the initial paper paragraphs to make sure that it mostly spans text from the introduction. %spans either abstract or introduction.
In the next step, we develop 10 different prompts that scientists frequently use during the writing process to refine their texts. To identify these prompts, we conducted an anonymous survey in which colleagues and students were asked to share the prompts they frequently use to improve their own writing.
The 10 final prompts ultimately include both more general requests (\textit{Improve the following}; \textit{Please polish the following text}) as well as prompts which ask for the improvement of specific text dimensions (\textit{Please improve the coherence of the text}; \textit{Refine grammar, tone, and readability}).
We then prompt GPT-3.5-turbo for each of the 3,000 original texts, randomly choosing one out of the 10 prompts which results in the final corpus. The final dataset (referred to as \textit{LLM}) consists of 6k paragraphs, 3k human-generated and 3k LLM-modified.

Note that over the full period covered by our naturalistic corpus, researchers may have used a range of different LLMs, particularly toward the end of \ttwo when newer models became available (e.g., GPT-4, Claude). However, %for the majority of the papers in our dataset, 
at the time of publication of most papers in our dataset, the most widely available and commonly used system was ChatGPT based on GPT-3.5. Therefore, we used GPT-3.5-turbo to generate paraphrases in our experiments.

\section{Characterizing Writing Changes via Lexical Choices}
\label{sec:lexical-choices}
%\section{Modeling Changes in Lexical Choices}
We begin by addressing \textbf{RQ1}: Are there core linguistic choices which changed between \tone and \ttwo in connection with LLM use? 
Focusing on the lexical level, we identify words with strongest changes in rates of use and then characterize them in terms of finer-grained patterns of semantic change.

\begin{figure*}
    \centering
    \includegraphics[width=\linewidth]{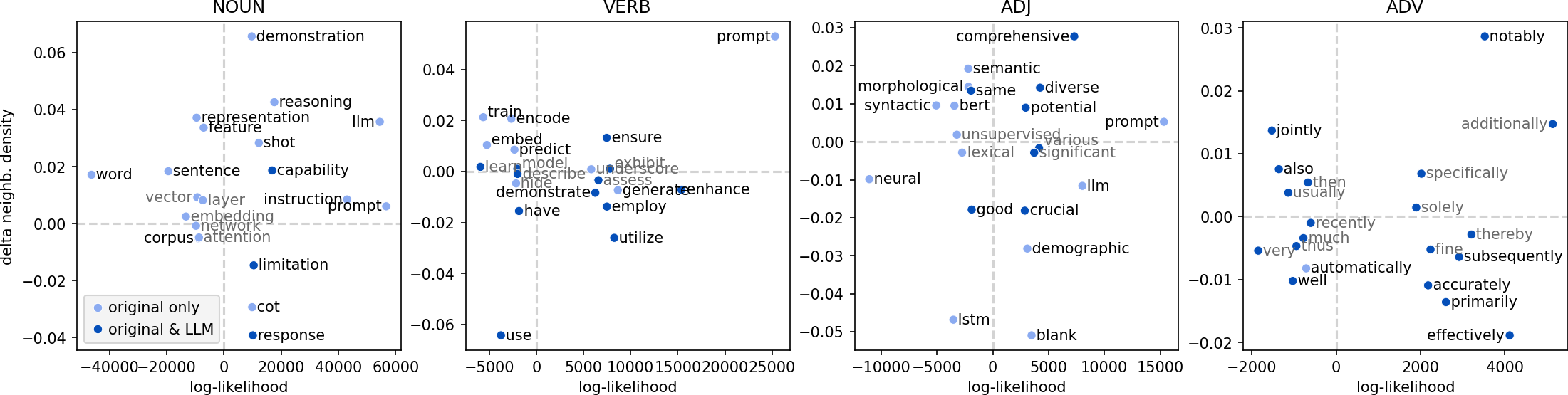}
    \caption{Target words with strongest differences in rates of use in \tone vs.\ \ttwo (top 10 per time period). X-axis: log-likelihood score, negative values indicate higher frequency in \tone. Y-axis: $\Delta ND$, higher values indicate an increase in neighborhood density over time (i.e., restriction of usage contexts); for grayed out targets, the difference in neighborhood density between \tone and \ttwo is not statistically significant. Color coding: dark blue targets also appear in the top 100 terms when contrasting original vs.\ LLM-paraphrased texts.}
    \label{fig:results-lexical}
\end{figure*}

\subsection{Experimental Setup}
\paragraph{Preprocessing}
% Starting from the text extracted from paper PDFs, we use spaCy\footnote{\url{https://spacy.io}} 3.8.4 (model \texttt{en\_core\_web\_sm}) to segment the text into sentences, which are then tokenized, lemmatized, and POS-tagged.
Starting from preprocessed paper text (cf.\ Section~\ref{sec:acl-antho-corpus}),
we examine content words (nouns, verbs, adjectives, and adverbs) in the shared vocabulary of \tone and \ttwo.
We lowercase all lemmas and retain those that are at least three characters long and contain only alphabetic characters.
% All lemmas are lowercased.
% The analysis is limited to content words (nouns, verbs, adjectives, and adverbs) in the shared vocabulary of \tone and \ttwo,
% with lemmas at least three characters long and containing only alphabetic characters.

\paragraph{Corpus statistics}
On the simplest level of analysis, we % use frequency-based statistics to 
quantify the extent to which a word's rate of use has changed between \tone and \ttwo.
To determine the strength of the change, we compute the log-likelihood score \citep{rayson-garside-2000-comparing}, which compares the observed frequencies of a given word in two corpora while accounting for the expected frequencies based on total corpus size.
To determine the directionality of the change, we calculate the frequency ratio of a word's frequency in \ttwo to its frequency in \tone (normalized per million tokens).
A value $<$1 is indicative of a term falling out of use over time, and vice versa.
% In addition to single lemmas, we also calculate these statistics for lemma 5-grams to check for changes in recurrent sequences (e.g., formulaic expressions).
We calculate these statistics both for the original corpus and the LLM dataset.

\paragraph{Type-level word embeddings}
To better understand broad semantic patterns behind lexical choices, we train type-level word embedding models.
We use word2vec \citep{mikolov2013efficient} in the gensim implementation \citep{rehurek2010gensim}, and set the algorithm to skip-gram with negative sampling, window size to 5, vector dimensions to 100, minimum frequency to 10, and other hyperparameters to default values.
We train separate models for \tone and \ttwo (original corpus), and run the process three times to account for randomness across word2vec runs \citep{pierrejean2018qualitative}. 
%We separately train word2vec models \citep{mikolov2013efficient} for \tone and \ttwo corpora, relying on the gensim implementation \citep{rehurek2010gensim}.
% We use skipgram algorithm with negative sampling, and set window size to 5, vector dimensions to 100, minimum frequency threshold to 10, and all other hyperparameters to default values.
% We train three models per corpus to account for stochastic differences across word2vec runs \citep{pierrejean2018qualitative}.

We characterize a word's usage in a given time period via its distributional neighborhood density,
which reflects semantic features such as polysemy and is therefore long-established in research on language change \citep{sagi-etal-2009-semantic}.
%This model-derived property efficiently captures core semantic features (e.g., polysemy) and as such is long-established in research on language change \citep{sagi-etal-2009-semantic}.
To calculate a target word's neighborhood density, we select its 100 nearest neighbors, and take the mean of the cosine similarity scores between the target and each neighbor.
We repeat the process for each of the three word2vec runs and take the average value.
We then calculate the change in neighborhood density for word $w$ between time periods \tone and \ttwo as
$\Delta ND(w) = ND(w_{t_2})-ND(w_{t_1})$.
% A higher value reflects an increase in neighborhood density over time, %(i.e., nearest neighbors becoming closer in vector space), 
% which is indicative of the target word being used in 
An increase in density is indicative of a restriction in usage contexts, typical of semantic specialization; %a trend typical of emergent technical terms.
a drop in density indicates diversification of usage contexts, which is typical of semantic generalization. %words gaining an additional sense.
We test the significance of changes in density using the Mann-Whitney--U test at 0.05 level.

\paragraph{Token-level word embeddings}
Connecting broad semantic trends to occurrence-level differences in usage, we implement an analysis using token-level embeddings.
For a given target word, we collect the sentences in which it occurs in the original corpus; these are capped at 1,000 per time period and randomly subsampled if necessary.
We then obtain contextualized embeddings of the target word in each sentence using ModernBERT \citep[base model with 22 layers, 768 dimensions, 149m parameters;][]{modernbert}.
%, which capture the surrounding context and can therefore distinguish between different usages.
% We use ModernBERT \citep{modernbert} in the base variant (22 layers, 768 dimensions, 149m parameters).
We feed the model with one sentence at a time, and retain the embedding corresponding to the target word in the last hidden state.
%If the target word is split into multiple subword tokens, we average over those.
%
%For each target word, all obtained embeddings are clustered together. %in order to identify main usage patterns across time.
The obtained embeddings for each target word are then clustered using using $k$-means, for which we rely on the scikit-learn implementation \citep{scikit-learn} and set $k$ to 8.
%We use the k-means algorithm in the scikit-learn implementation \citep{scikit-learn} and set $k$ to 8.
For each cluster, we calculate the proportion of examples from \tone and \ttwo, and draw on this information to identify stable vs.\ shifting usages. %which are stable vs.\ became more or less prominent over time.

\subsection{Results}

On the most general level, we inspect changes in rates of use for the whole vocabulary based on the distribution of log-likelihood scores.
This change is significant%
\footnote{Based on the critical log-likelihood value of 15.13 recommended by \citet{rayson2004extending}.}
for 13\% (9,353 out of 71,993) words in the shared vocabulary as defined above.
% All words: 71993
% Signif LL: 9353 12.99
% Across parts of speech, this rate is highest for ...
% ll	ll_signif	perc
% pos			
% ADJ	19168	2305	12.03
% ADV	3478	428	12.31
% NOUN	37801	5316	14.06
% VERB	11546	1304	11.29
The significant changes are near-evenly distributed between drops (45\%) and increases (55\%) in frequency,
matching the intuition that the obsolescence of most words is paralleled by a rise in prominence of their functional equivalents.
%most words which come out of use are in parallel substituted by their functional equivalents.
We further analyze the words with significant changes in rates of use by correlating their log-likelihood score (to which we assign a negative value for words whose frequency drops over time) and the change in neighborhood density $\Delta ND$.
We find a negative correlation (Spearman's $\rho=-0.26$, $p\ll0.01$)
suggesting that increased rate of use is associated with a lowering of neighborhood density, which typically reflects semantic generalization. % (shown by a lowering of neighborhood density). %, which in turn reflects a broader range of usage contexts);
However, the limited strength of the correlation indicates that the process does not apply to the whole vocabulary.%, which is likely better understood by analyzing the complementary information conveyed by the two scores.

%We adopt such a complementary perspective to more closely investigate the strongest changes.
We now shift from vocabulary-level trends to the most relevant individual words.
For each part-of-speech, we retain the words with the 10 highest positive and negative log-likelihood scores, respectively capturing the strongest increases and drops in rate of use over time.
We plot these words in Figure~\ref{fig:results-lexical}, with log-likelihood on the x-axis and change in neighborhood density $\Delta ND$ on the y-axis.

Changes in rates of use (x-axis) often reflect patterns of lexical replacement. % affecting both specialized topics and general-language expressions.
Some reflect shifts in specialized topics, such as
%For example, in the nouns plot, 
technical terms referring to core machine learning concepts %, \textit{vector}) 
dropping out of use (e.g., \textit{embedding}), and those related to more recent approaches %(e.g., \textit{LLM}, \textit{prompt}) 
increasing their rate of use (e.g., \textit{prompt}).
% We find a similar shift in verbs, moving away from terms such as \textit{train} % and \textit{embed} and 
% towards those like \textit{generate}. %and \textit{prompt}
Other patterns capture stylistic rather than terminological differences:
general-language expressions which are semantically broad and stylistically neutral tend to fall out of use, while more formal and specialized words gain prominence.
This trend is especially visible for verbs (e.g., \textit{use} vs.\ \textit{utilize}), adjectives (e.g., \textit{good} vs.\ \textit{comprehensive}), and adverbs (e.g., \textit{then} vs.\ \textit{subsequently}).
% Across the verb, adjective, and adverb plots, we note terms which are both semantically broad and stylistically neutral (e.g., \textit{use}, \textit{good}, \textit{then}) falling out of use, 
% %(e.g., \textit{use}, \textit{have}; \textit{good}, \textit{same}; \textit{well}, \textit{then}) 
% with more formal terms rising in frequency (e.g., \textit{utilize}, \textit{comprehensive}, \textit{subsequently}).

But can these changes be attributed to LLM-assisted writing? %, or are they driven by more trivial factors such as a shift in dominant topics within the NLP community?
%To address this question, we use 
We inspect the log-likelihood scores calculated on our LLM dataset, comparing original \tone texts and their LLM-assisted paraphrases. %; by design, this setting keeps the topic fixed.
We select the words with 100 highest positive and negative log-likelihood scores (per part of speech).
The overlap between this set and the top-10 sets from the original corpus is shown in Figure~\ref{fig:results-lexical} using dark blue markers.
%Figure~\ref{fig:results-lexical} uses darker blue markers to show the most distinctive words from the original corpus which \textit{also} feature among the most distinctive words from the LLM corpus.
The overlap is almost entirely limited to general-language and not technical terms.
Given the topic-controlled nature of the LLM corpus, this finding indicates that (i)~the changes in the original corpus reflect two parallel trends: a topical and a stylistic shift; and (ii)~that the stylistic shift can be attributed to LLM use.

We now analyze the semantic change mechanisms associated with different rates of use, as measured by the change in neighborhood density $\Delta ND$ (y-axis) and by the temporal distribution over clusters of token-level embeddings.
% We now turn to the characterization of semantic change by neighborhood density.
% Across parts of speech, subgroups of target expressions do not show statistically significant differences in neighborhood density (shown as grayed out target words (e.g., \textit{describe}, \textit{significant}).
% Perhaps more importantly, general-language terms also differ in terms of the semantic mechanisms connected to their differences in rates of use.
As an example, we focus on the verbs \textit{ensure} and \textit{utilize}: they both show an increase in frequency over time, but differ in semantic mechanisms.
The verb \textit{ensure} shows %weak (but statistically significant) 
an increase in neighborhood density, which is indicative of semantic specialization.
This trend is supported by clustered examples like the following:
\begin{enumerate}[(1), nolistsep]
    \item Lastly, we aim to measure the semantic similarity between generated questions to \textbf{ensure} that the questions assess the same content. \label{ex-ensure-soas}
    \item To \textbf{ensure} the high quality of the annotation procedure, we manually annotated a set of 200 control tasks. \label{ex-ensure-quality}
\end{enumerate}

Example~\ref{ex-ensure-soas} is from a cluster dominated by older data (65\% \tone) where \textit{ensure} conveys a broad sense of finality across diverse contexts.
Example~\ref{ex-ensure-quality} comes from a more recent cluster (57\% \ttwo).
It is representative of the more specific meaning ‘to guarantee', attested in a restricted set of transitive contexts usually referring to the quality of a scientific artifact.

In contrast, \textit{utilize} shows a slight expansion of distributional neighborhood, typical of semantic generalization which is also borne out by these examples:
\begin{enumerate}[(1), resume, nolistsep]
    \item The latter aims to \textbf{utilize} the multi-level interests to enhance both conversation and recommendation tasks when users chat with system.\label{ex-utilize-specific}  %over time. 
    \item Finally, we \textbf{utilize} a ridge regression classifier to obtain final classification results. \label{ex-utilize-generic} 
\end{enumerate}

Example~\ref{ex-utilize-specific} is from an older cluster (60\% \tone) and conveys the specific meaning of ‘use to the fullest potential'.
Example~\ref{ex-utilize-generic} comes from a cluster with more recent data (56\% \ttwo) where \textit{utilize} is attested with the broad meaning ‘make use of'.%

Summarizing, we identified a substantial set of the vocabulary with a shift in the rate of use since the introduction of ChatGPT.
As further validation, in Appendix~\ref{app:lexical} we also present complete corpus statistics for the strongest changes in rates of use, extend the same analysis to multi-word sequences, and analyze further clustering examples. 
We consistently find that some changes are due to topical shifts within the scientific community, while others are more clearly attributable to the adoption of LLMs. %, as further confirmed by the comparison with our synthetic corpus; in general, "more complex" words becoming more common.
%The changes do not merely affect the rate at which the words are used, but also their semantic properties as reflected by usage contexts; 
%These changes are not only reflected in a different frequency, but also in shifts in the dominant senses, 
The changes are further reflected in shifting sense distributions, leading to semantic generalization in some cases and specialization in others.
%We now broaden the perspective to include further types of linguistic properties.

\section{Predicting Time Periods from Complex Linguistic Features}
In the following section, we address \textbf{RQ2}: are there systematic stylistic differences between \tone and \ttwo that could be attributed to the use of LLMs as writing assistants? 
\begin{figure*}
    \centering
    \includegraphics[width=\linewidth]{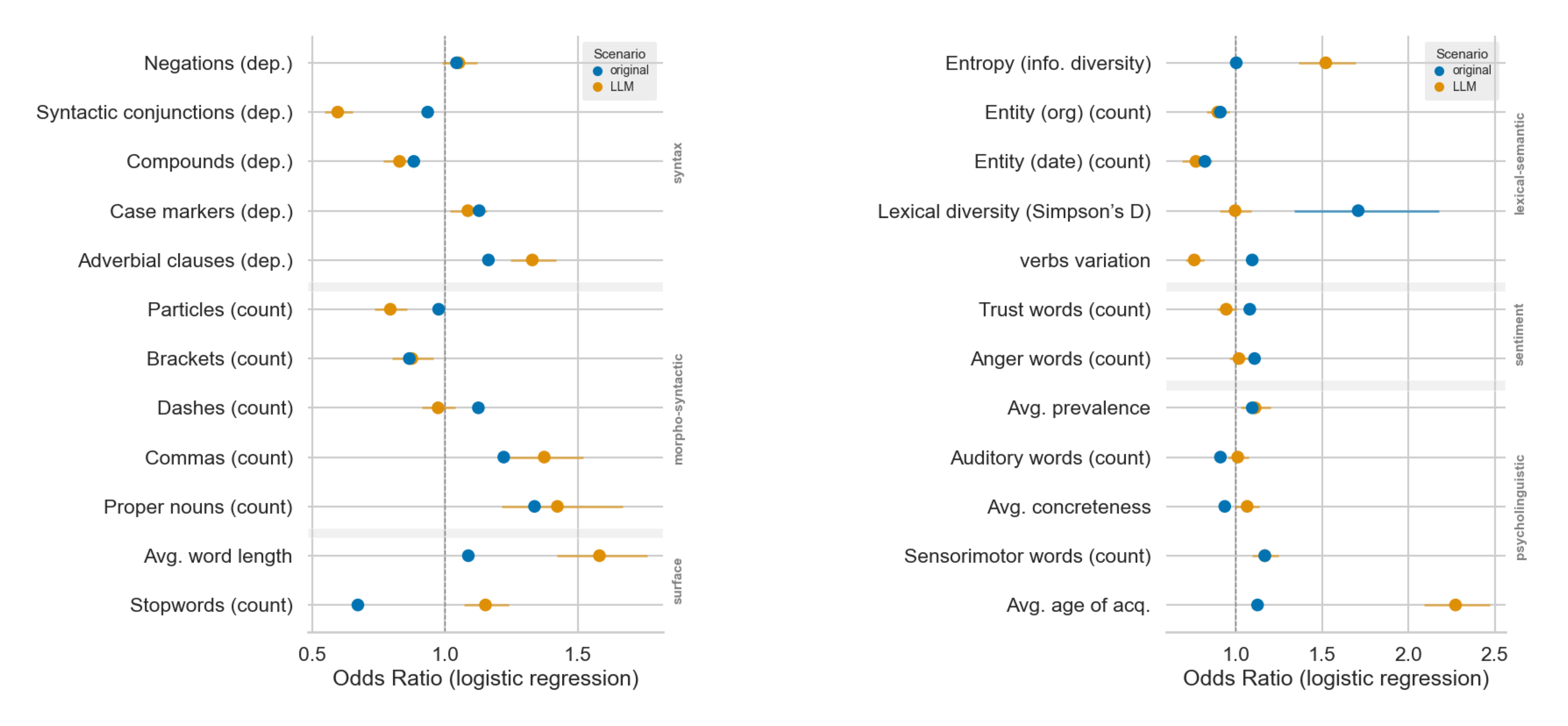}
    \caption{Comparison of odds ratios for linguistic features between the original and LLM-paraphrased datasets. Positive odds ratios indicate that higher feature values are more characteristic of the post-GPT / LLM-paraphrased texts. Horizontal bars represent 95\% CIs (approximated with the Wald method for original). The vertical dashed line at 1 marks the point of no effect.}
    \label{fig:coeff}
\end{figure*}

\subsection{Experimental Setup}
We apply logistic regression as an analysis tool to find the linguistic markers that significantly contribute to explaining the outcome (a binary variable representing whether a text is from \tone or \ttwo), after accounting for other relevant feature groups.
The advantage of this approach is that it allows us to model the relationship between each feature and the target variable: positive odds indicate that higher values of a feature are characteristic of \ttwo (human + LLM assistant), negative odds indicate that they are more characteristic of \tone (humans only). We can also reveal which feature groups have the greatest impact. 

Since many factors may influence stylistic change in our real-world data (original dataset), we repeat the same analysis in the controlled, synthetic setting (LLM dataset) to isolate stylistic effects specifically linked to LLM use.

\paragraph{Preprocessing}
We first extract around 1,000 linguistic features from six different features groups with \texttt{elfen} \cite{maurer-2025-elfen} and \texttt{LFTK} \cite{lee-lee-2023-lftk} for all paragraphs from \tone and \ttwo in the original dataset, as well as all paragraphs in the LLM dataset. The features cover surface features (e.g., word and sentence length), morpho-syntactic features (e.g., occurrences of specific morphological constructions or part-of-speech tags), syntactic features (e.g., dependency relations and the complexity of dependency trees), psycholinguistic features (e.g., acquisition norms or average concreteness of words), lexical-semantic features (e.g., measures of lexical diversity or occurrences of specific named entities), and sentiment features (polarity and emotion-related words). All features are  standardized and scaled. 

\paragraph{Methodology}
As a first step, we apply a filtering method to select a meaningful set of features as predictors: We keep only features with variance $>$0 and perform a correlation analysis, retaining features with correlation $<$0.7 and selecting one representative feature per highly correlated group. This yields 145 initial features.

We then perform stability selection using logistic regression with elastic net regularization on a balanced subset of 200k paragraphs from the \textit{original} dataset, applying 5-fold inner and 10-fold outer cross-validation to assess robustness. 
\camerareadytext{In each fold, we record whether a feature is selected (coefficient > 0), determine its effect direction, and calculate the change in odds ratio. We then apply bootstrapping to estimate confidence intervals as an approximation of statistical significance.}
For our final selection we remove all features that (a)~are not selected in all folds (a feature is selected when coefficient $>$0), (b)~show inconsistent effects, (c)~show less than $\pm$5\% change in odds, and (d)~were not significant (we bootstrapped CIs to approximate significance).
This leaves 45 robust features. To further reduce the number, we rank them by absolute change in odds and select the top features per feature group, resulting in 24 final features.
Finally, we refit the logistic regression on the full dataset for unbiased estimates. For the synthetic scenario, we use a generalized mixed-effects model with paper ID as a random effect to account for data dependencies.
\camerareadytext{
For final, unbiased estimates, we use the full dataset and fit a logistic regression model with the 24 selected features on the real-world data. In the synthetic scenario, data points are not independent since we used a text from a unique paper and paraphrased it. To account for the grouping structure we use a generalized mixed-effects model with paper ID as a random effect.}

\subsection{Results}
The regression model on the original dataset achieves a mean AUC of 0.65 $\pm$0.002 and a pseudo $R^2$ of 5.6\% (McFadden) across the 10-fold cross-validation, which means that it explains not a large but meaningful amount of variance. Using the same features as fixed effects in the mixed effects model explains 29\% of the variance, which is a substantial portion and confirms that LLM-modified texts exhibit particular stylistic patterns that are characterized well by the linguistic features.

Figure \ref{fig:coeff} visualizes the direction and strength of association between each feature and the outcome variable (\tone vs.\ \ttwo or human vs.\ GPT-paraphrased), grouped by the different feature groups. Points to the right of the 1.0 decision boundary indicate that higher feature values are associated with LLM use, while points to the left correspond to human-written texts. When blue and orange points appear on the same side, the pattern is consistent across both the original and LLM datasets.

LLM use is characterized by longer and more complex words (higher average word length and age of acquisition), confirming the findings from the previous section.
They show greater entropy while still containing many familiar, high-prevalence terms. In contrast, human-written texts use more stopwords, suggesting that LLM outputs are more semantically dense. Stopword use, however, presents a contrasting pattern: while it shows a strong positive association in the original corpus, the relationship is reversed in the LLM-modified texts.
Human-written texts tend to be more varied, showing higher lexical diversity (Simpson’s D) and greater verb variability, although this pattern is not entirely consistent.

Overall, the findings present a mixed picture of lexical variation and linguistic complexity: LLMs appear to produce more complex syntactic constructions and lexically dense content, yet with less lexical variety and a preference for familiar vocabulary. Whether this results in improved or reduced clarity will be examined in the following section.

In terms of syntactic and morpho-syntactic structures, there are distinct patterns between human and LLM-modified texts. LLMs tend to use more negations and adverbial clauses, indicating a preference for more elaborated or qualified sentence structures with additional modifiers. Human-written texts, in contrast, are more strongly associated with conjunctions -- a construction often simplified or replaced with more complex connectives by LLMs -- and contain more compounds.

We observe clear differences in punctuation use: human-written texts more often include brackets, whereas LLM-modified texts introduce more commas and possibly dashes.
LLM-improved texts also contain more proper nouns, with the exception of organization entities, which occur more frequently in human-written texts. Similarly, date entities are more strongly associated with human-written texts, as LLMs are less likely to generate references and often remove them when prompted to improve human originals.

Sentiment patterns show a mixed picture: LLMs use more anger-related words, while original texts contain more trust-related words expressing confidence or credibility (e.g., \textit{confirm}, \textit{promise}, \textit{support}) -- although this trend is not confirmed in the paraphrased corpus. Another interesting pattern is the higher frequency of sensorimotor words in LLM-modified texts (e.g., \textit{enhance}, \textit{highlight}), possibly because LLMs were trained to make writing more dynamic and engaging. This could point to a stylistic bias learned from human feedback and exposure to polished text.

Finally, we investigate the relative importance of each feature. Looking at the most important features, we find that 
dependency features (e.g., adverbial clauses) and punctuation have a major influence on both models, indicating a generalizable, strong pattern in terms of the difference between human and LLM-generated texts. Another strong feature in both datasets is the more frequent use of proper nouns in LLM-modified texts.  The largest difference between the two datasets lies in their key predictive features: in the original corpus, variance is primarily explained by stopword use, whereas in the LLM dataset, it is driven by the average age of acquisition.

\section{Measuring Reading Experience}
While prior research has examined lexical and stylistic differences between human- and LLM-generated texts, little is known about how readers perceive these changes. To address this gap, we conducted a pilot study to assess human perception of human-only and LLM-modified texts (\textbf{RQ3}).

To gain insights on that question, we look at four broader dimensions of reading experience: \textit{clarity} measures whether a text is understandable and communicates the content clearly, \textit{authenticity} captures to what extent the reader feels connected to the author and perceives the author as being genuine in their writing, \textit{trustworthiness} captures the extent to which a text presents its arguments clearly, reliably, and credibly, while \textit{excitement} assesses whether the reading experience is engaging and stimulates the reader’s interest.

\subsection{Annotation Setup}
\paragraph{Study design}
To find out whether readers prefer human-written or LLM-improved texts and along which dimensions, we designed the annotation study as a pairwise comparison task. Given a pair of texts, text A and text B, annotators had to rate which one aligns more strongly with a certain statement. 
We measure each dimension with two statements, for example, \textit{I read this text smoothly and fluently} to measure the dimension of \textit{clarity}. Raters indicated their preference on a four-point Likert scale (strongly A, slightly A, slightly B, strongly B). We asked 20 different domain experts (NLP researchers) with varying backgrounds and levels of seniority to annotate 20 text-pairs. For each pair we collected annotations from two raters, resulting in 200 annotated instances. Appendix \ref{app:annotation} provides the guidelines and detailed questionnaire. 

\paragraph{Annotation data}
We use a subset of the LLM dataset to (a) control for topic effects and (b) ensure clear gold labels. To select good candidates, each original text and its paraphrase were converted into feature vectors capturing style (linguistic features) and semantics \citep[using SBERT embeddings;][model \texttt{paraphrase-multilingual-mpnet-base-v2}]{reimers2019sentencebert}. We calculated pairwise differences and ranked pairs by average distance. We then manually selected 200 high-quality pairs from the top 300, removing noisy instances, LLM artifacts, and formatting indicators.

\subsection{Results}
\begin{figure}
    \centering
    \includegraphics[width=\linewidth]{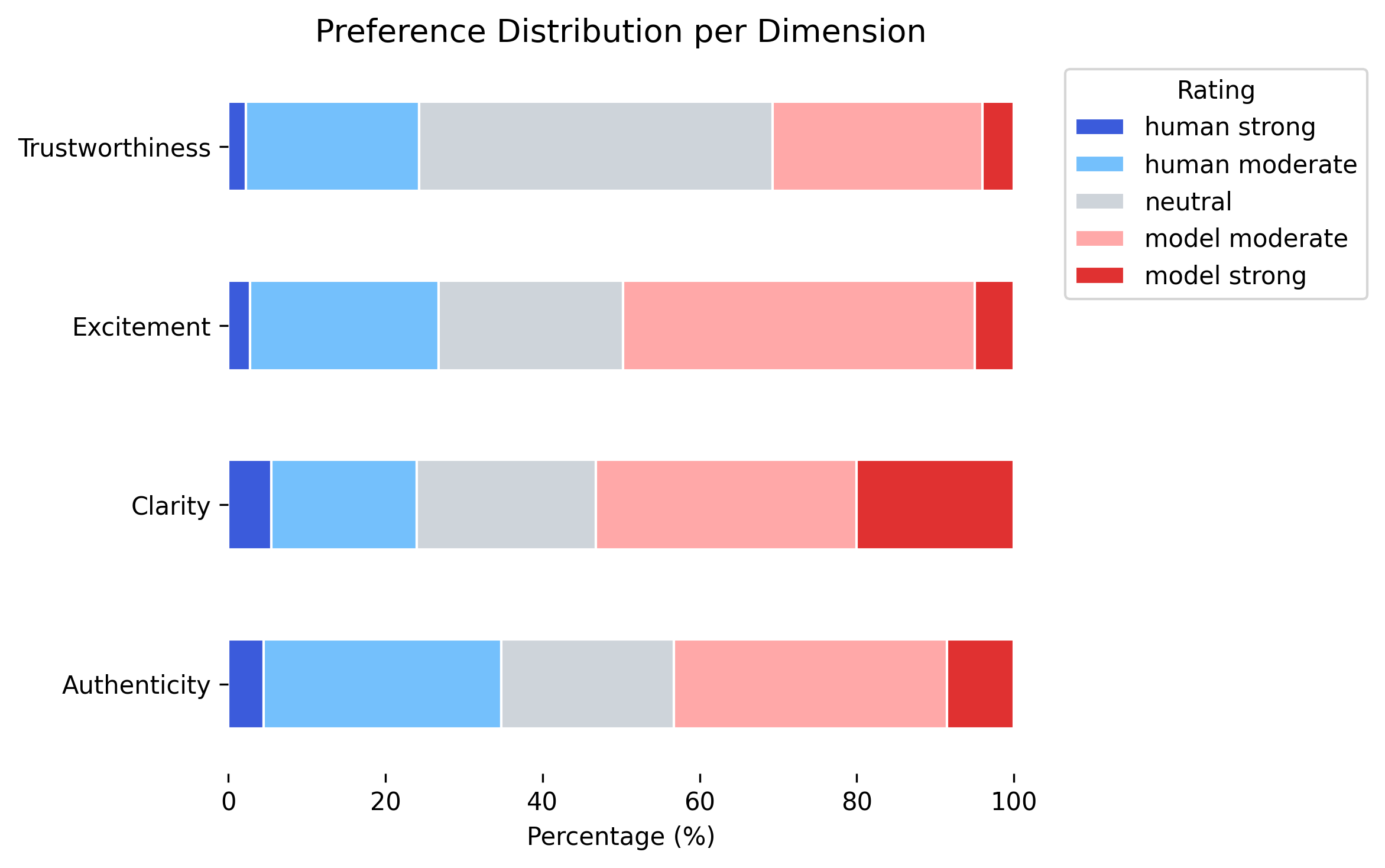}
    \caption{Distribution over Preferences in Human Raters. Higher red portion indicates a larger preference for LLM-improved texts, higher blue portion for the human originals.}
    \label{fig:preferences}
\end{figure}

\paragraph{Quantitative ratings}
Figure \ref{fig:preferences} shows the overall preference distribution of human raters in our annotation study. We can see that raters tend to prefer LLM-improved texts, especially on the dimensions of clarity and excitement. However, there is also a substantial portion of texts where the human original was preferred, which shows that prompting an LLM to improve a scientific text does not always lead to the desired effect on the reader side. Additionally, a considerable number of texts showed no clear preference for either version. 

For each dimension, we measure whether the ratings significantly deviate from 0 (neutral) and quantify the effect size. Table \ref{tab:singpreference} shows that the strongest preference for LLM-improved texts can be observed in \textit{clarity} and \textit{excitement}. The clear preference for LLM-improved texts in \textit{clarity} is unsurprising, as this aspect is often explicitly mentioned in the prompts. Interestingly, even for dimensions not directly prompted, such as \textit{authenticity}, LLM texts are still preferred more often, though the effect size is small. 

When comparing individual annotators, we also find notable differences. 
Some clearly favor human-written texts on the dimensions of \textit{trustworthiness} and \textit{authenticity}, while others strongly prefer the LLM-modified versions, indicating that these two dimensions are perceived more subjectively.
\begin{table}[t]
\centering
\resizebox{\linewidth}{!}{%
\begin{tabular}{lrrrrr}
\toprule
\textbf{Measure} & \textbf{Mean} & \textbf{t}  & \textbf{p (Wilcoxon)} & \textbf{Cohen's $d$} \\
\midrule
Clarity        & -0.44 & -7.53 &  \textbf{0.000} & -0.38 \\
Authenticity   & -0.12 & -2.32 &  \emph{0.019}   & -0.12 \\
Trustworthiness & -0.08 & -1.93 & 0.054          & -0.10 \\
Excitement     & -0.25 & -5.22 &  \textbf{0.000} & -0.26 \\
\bottomrule
\end{tabular}}
\caption{Model preference scores (range: -2 to 2; 0 = no preference). Negative values indicate preference for LLM-paraphrased texts, positive values for human originals. LLM paraphrases are favored for clarity and excitement (strong direction, moderate effect size). Significance assessed using the Wilcoxon rank test.}
\label{tab:singpreference}
\end{table}

\paragraph{Qualitative remarks}
We conducted qualitative interviews with five participants after the annotation study.
They all described the annotation task as challenging; with the exception of clarity-related ratings, they reported difficulties due to the subjective nature of most questions.
They further noted lower confidence particularly for the ratings on the intermediate points of the scale, underscoring the need for a nuanced reading of the results.

%While the task instructions purposefully did not describe how the rated texts were obtained, all interviewed participants assumed that LLM use was involved.
Most participants made intuitive assumptions about which text in the pair was LLM-based, but they also noted a limited degree of certainty. % and changes of heart.
They collectively reported a wide range of properties leading them to think a text was LLM-assisted: limited variability in lexical choices and sentence length; the amount of references and abbreviations; formatting, e.g., punctuation and bullet lists.
The diversity of these strategies and their limited accuracy highlights possible interactions between personal attitudes and reading experience. 
A mismatch is further supported by generally negative expressed attitudes towards LLM writing (in this subgroup of participants) vs.\ overall positive collected ratings.

% - all participants found the annotation task challenging
% - most thought that clarity-related questions were straightforward, but had a harder time deciding on the more subjective ones; cautioned regarding their own confidence
% - intuition about LLMs, but not always sure
% - qualitatively described (assumed) LLM text as verbose, polished but not conveying the point, limited in variation (regarding lexical choices but also sentence structure)
% - generally took into consideration the fact that the texts were being read out of the context of the full paper, without clear-cut trends which would question the reliability of the data collection
% They collectively reported a wide range of strategies to recognize LLM writing: formatting, e.g. punctuation and bullet lists; the amount of references and abbreviations; limited variability in lexical choices and sentence length.
% The diversity of these strategies and their unreliable (sometimes outright incorrect) link with LLM writing highlights interesting interactions between well-established personal attitudes and actual reading experience. 
% This is further supported by the generally negative expressed attitudes towards LLM writing (at least in this subgroup of participants) vs. generally more positive ratings.

\section{Conclusion}
In this work, we investigated the influence of LLMs on writing style in scientific communication, specifically in the NLP domain.
We relied on two data scenarios: a naturalistic corpus of over 37,000 scientific papers published in the two years preceding vs.\ following the release of ChatGPT,
and a synthetic dataset of 3,000 human-written passages and their LLM-generated improvements.

With regard to word usage, we found that both word frequency and the contexts in which these are used have changed significantly, indicating semantic specialization in some cases and generalization in others.
We also found specific stylistic features that distinguish LLM-modified texts from human texts. For example, LLM-modified texts more frequently contain certain syntactic constructions, more complex and longer words and a lower lexical diversity.
Crucially, trends in word usage as well as stylistic features are broadly consistent across the naturalistic and synthetic data scenarios, indicating that many observed shifts in writing practices can be attributed to growing LLM use.
Finally, in a pilot study, we measured the impact of these changes in writing style on the reading experience. The results indicate that LLM-improved texts are perceived as more understandable and exciting.

Our findings encourage further research along several dimensions. These include a larger-scale annotation study to verify the generalizability of the results; exploring prompt- or model-specific characteristics of writing style and word usage; and deploying the identified linguistic features in applied scenarios such as detecting AI-generated content.

\section{Limitations}
We note several limitations of our work.
First, our analysis is framed as a binary comparison of language use across two time periods. This approach has important practical benefits and is underpinned by clearly stated assumptions (e.g., uncertainty regarding the precise degree of LLM use in the second time period).
However, a finer-grained comparison of smaller time slices could provide a clearer picture regarding the development patterns of LLM-supported stylistic choices.

We also acknowledge that, prior to the release of ChatGPT, other AI-assisted writing tools such as Grammarly were already available and may have been used during the first time period, potentially influencing some publications. However, these tools primarily focus on grammar correction and minor stylistic suggestions rather than generating substantial new text or paraphrases. For this reason, their potential impact on the linguistic patterns we analyze is likely more limited compared to modern large language models. A larger empirical study comparing the stylistic differences between traditional writing tools and LLMs as writing assistants would be a valuable direction for future work.

More generally, our study is limited to a single scientific domain (Natural Language Processing) and one European language (English).
Since academic writing conventions are highly specific to disciplines as well as languages, replicating this analysis on more diverse datasets would help assess the generalizability of the trends we report.

Finally, we ran a pilot annotation study with 20 participants, collecting two judgments per instance.
A larger-scale setup with more annotations per instance would provide more robust results.

\section{Acknowledgments}
We are grateful to our 20 volunteer annotators for the time and effort they put into this study.
% We are grateful to our 20 volunteer annotators: 
% Ana Barić, Hongyu Chen, Esra Dönmez, Anne Eichel, Agnieszka Faleńska, Lynn Greschner, Maximilian Maurer, Chris Jenkins, Allison Keith, Nora Ketschik, Urban Knupleš, Sinan Kurtyigit, Laura Majer, Aleksandra Miletić, Prisca Piccirilli, Vigneshwaran Shankaran, Tarun Tater, Nina Vikhrova, Franziska Weeber, and Amelie Wührl.
Filip Miletić was supported by DFG Research Grant SCHU 2580/5-2 (\textit{Computational Models of Semantic Variation in Multi-Word Expressions across Speakers and Languages}).

%\clearpage
\section{Bibliographical References}
\label{sec:reference}
\bibliographystyle{lrec2026-natbib}
\bibliography{references_fm,lrec2026-example}

\begin{thebibliography}{46}
\expandafter\ifx\csname natexlab\endcsname\relax\def\natexlab#1{#1}\fi

\bibitem[{Akinwande et~al.(2024)Akinwande, Adeliyi, and Yussuph}]{Akinwande_2024}
Mayowa Akinwande, Oluwaseyi Adeliyi, and Toyyibat Yussuph. 2024.
\newblock \href {https://doi.org/10.5121/ijci.2024.130408} {Decoding ai and human authorship: Nuances revealed through nlp and statistical analysis}.
\newblock \emph{International Journal on Cybernetics \& Informatics}, 13(4):85–103.

\bibitem[{Chen et~al.(2023)Chen, Kang, Zhai, Li, Singh, and Raj}]{chen-etal-2023-token}
Yutian Chen, Hao Kang, Vivian Zhai, Liangze Li, Rita Singh, and Bhiksha Raj. 2023.
\newblock \href {https://doi.org/10.18653/v1/2023.emnlp-main.810} {Token prediction as implicit classification to identify {LLM}-generated text}.
\newblock In \emph{Proceedings of the 2023 Conference on Empirical Methods in Natural Language Processing}, pages 13112--13120, Singapore. Association for Computational Linguistics.

\bibitem[{Desaire et~al.(2023)Desaire, Chua, Isom, Jarosova, and Hua}]{DESAIRE23}
Heather Desaire, Aleesa~E. Chua, Madeline Isom, Romana Jarosova, and David Hua. 2023.
\newblock \href {https://doi.org/https://doi.org/10.1016/j.xcrp.2023.101426} {Distinguishing academic science writing from humans or chatgpt with over 99\% accuracy using off-the-shelf machine learning tools}.
\newblock \emph{Cell Reports Physical Science}, 4(6):101426.

\bibitem[{Doru et~al.(2025)Doru, Maier, Busse, L\"{u}cke, Sch\"{o}nhoff, Enax-Krumova, Hessler, Berger, and Tokic}]{Doru2025}
Berin Doru, Christoph Maier, Johanna~Sophie Busse, Thomas L\"{u}cke, Judith Sch\"{o}nhoff, Elena Enax-Krumova, Steffen Hessler, Maria Berger, and Marianne Tokic. 2025.
\newblock \href {https://doi.org/10.2196/62779} {Detecting artificial intelligence–generated versus human-written medical student essays: Semirandomized controlled study}.
\newblock \emph{JMIR Medical Education}, 11:e62779.

\bibitem[{Dugan et~al.(2024)Dugan, Hwang, Trhl{\'i}k, Zhu, Ludan, Xu, Ippolito, and Callison-Burch}]{dugan-etal-2024-raid}
Liam Dugan, Alyssa Hwang, Filip Trhl{\'i}k, Andrew Zhu, Josh~Magnus Ludan, Hainiu Xu, Daphne Ippolito, and Chris Callison-Burch. 2024.
\newblock \href {https://doi.org/10.18653/v1/2024.acl-long.674} {{RAID}: A shared benchmark for robust evaluation of machine-generated text detectors}.
\newblock In \emph{Proceedings of the 62nd Annual Meeting of the Association for Computational Linguistics (Volume 1: Long Papers)}, pages 12463--12492, Bangkok, Thailand. Association for Computational Linguistics.

\bibitem[{Gao et~al.(2023)Gao, Howard, Markov, Dyer, Ramesh, Luo, and Pearson}]{Gao2023}
Catherine~A. Gao, Frederick~M. Howard, Nikolay~S. Markov, Emma~C. Dyer, Siddhi Ramesh, Yuan Luo, and Alexander~T. Pearson. 2023.
\newblock \href {https://doi.org/10.1038/s41746-023-00819-6} {Comparing scientific abstracts generated by {ChatGPT} to real abstracts with detectors and blinded human reviewers}.
\newblock \emph{npj Digital Medicine}, 6(1).

\bibitem[{Gray(2024)}]{Gray2024ChatGPTE}
Andrew Gray. 2024.
\newblock \href {https://arxiv.org/abs/2403.16887} {{ChatGPT} "contamination": estimating the prevalence of {LLMs} in the scholarly literature}.
\newblock \emph{arXiv}, abs/2403.16887.

\bibitem[{Grootendorst(2022)}]{grootendorst2022bertopic}
Maarten Grootendorst. 2022.
\newblock \href {https://arxiv.org/abs/2203.05794} {{BERTopic}: Neural topic modeling with a class-based {TF-IDF} procedure}.
\newblock \emph{arXiv}, abs/2203.05794.

\bibitem[{Guggilla et~al.(2025)Guggilla, Roy, Chavan, Rahman, and Bowen}]{Guggilla2025AIGT}
Chinnappa Guggilla, Budhaditya Roy, Trupti Chavan, Abdul Rahman, and Edward Bowen. 2025.
\newblock \href {https://arxiv.org/abs/2507.05157} {{AI} generated text detection using instruction fine-tuned large language and transformer-based models}.
\newblock \emph{arXiv}, abs/2507.05157.

\bibitem[{Guo et~al.(2023)Guo, Zhang, Wang, Jiang, Nie, Ding, Yue, and Wu}]{HC3}
Biyang Guo, Xin Zhang, Ziyuan Wang, Minqi Jiang, Jinran Nie, Yuxuan Ding, Jianwei Yue, and Yupeng Wu. 2023.
\newblock \href {https://doi.org/10.48550/ARXIV.2301.07597} {How close is {ChatGPT} to human experts? {C}omparison corpus, evaluation, and detection}.
\newblock \emph{arXiv}, abs/2301.07597.

\bibitem[{Hakam et~al.(2024)Hakam, Prill, Korte, Lovrekovi{\'c}, Ostoji{\'c}, Ramadanov, and Muehlensiepen}]{Hakam2024-ql}
Hassan~Tarek Hakam, Robert Prill, Lisa Korte, Bruno Lovrekovi{\'c}, Marko Ostoji{\'c}, Nikolai Ramadanov, and Felix Muehlensiepen. 2024.
\newblock \href {https://doi.org/10.2196/52164} {Human-written vs {AI-generated} texts in orthopedic academic literature: Comparative qualitative analysis}.
\newblock \emph{JMIR Formative Research}, 8:e52164.

\bibitem[{Hamed and Wu(2023)}]{hamedWu}
Ahmed~Abdeen Hamed and Xindong Wu. 2023.
\newblock \href {https://doi.org/10.48550/ARXIV.2308.11767} {Improving detection of {ChatGPT}-generated fake science using real publication text: {I}ntroducing {xFakeBibs} a supervised-learning network algorithm}.
\newblock \emph{arXiv}, abs/2308.11767.

\bibitem[{Kobak et~al.(2025)Kobak, González-Márquez, Horvát, and Lause}]{Kobak2025}
Dmitry Kobak, Rita González-Márquez, Emőke-Ágnes Horvát, and Jan Lause. 2025.
\newblock \href {https://doi.org/10.1126/sciadv.adt3813} {Delving into {LLM}-assisted writing in biomedical publications through excess vocabulary}.
\newblock \emph{Science Advances}, 11(27).

\bibitem[{Koller et~al.(2024)Koller, Beam, Manrai, Ashley, Liu, Gichoya, Holmes, Zou, Dagan, Wong, Blumenthal, and Kohane}]{Koller2024}
Daphne Koller, Andrew Beam, Arjun Manrai, Euan Ashley, Xiaoxuan Liu, Judy Gichoya, Chris Holmes, James Zou, Noa Dagan, Tien~Y. Wong, David Blumenthal, and Isaac Kohane. 2024.
\newblock \href {https://doi.org/10.1056/aie2300128} {Why we support and encourage the use of large language models in {NEJM AI} submissions}.
\newblock \emph{NEJM AI}, 1(1).

\bibitem[{Lee and Lee(2023)}]{lee-lee-2023-lftk}
Bruce~W. Lee and Jason Lee. 2023.
\newblock \href {https://aclanthology.org/2023.bea-1.1} {{LFTK}: Handcrafted features in computational linguistics}.
\newblock In \emph{Proceedings of the 18th Workshop on Innovative Use of NLP for Building Educational Applications (BEA 2023)}, pages 1--19, Toronto, Canada. Association for Computational Linguistics.

\bibitem[{Lee et~al.(2022)Lee, Liang, and Yang}]{coauthor}
Mina Lee, Percy Liang, and Qian Yang. 2022.
\newblock \href {https://doi.org/10.1145/3491102.3502030} {{CoAuthor}: Designing a human-{AI} collaborative writing dataset for exploring language model capabilities}.
\newblock In \emph{Proceedings of the 2022 CHI Conference on Human Factors in Computing Systems}, CHI '22, New York, NY, USA. Association for Computing Machinery.

\bibitem[{Li et~al.(2024)Li, Li, Cui, Bi, Wang, Wang, Yang, Shi, and Zhang}]{li-etal-2024-mage}
Yafu Li, Qintong Li, Leyang Cui, Wei Bi, Zhilin Wang, Longyue Wang, Linyi Yang, Shuming Shi, and Yue Zhang. 2024.
\newblock \href {https://doi.org/10.18653/v1/2024.acl-long.3} {{MAGE}: Machine-generated text detection in the wild}.
\newblock In \emph{Proceedings of the 62nd Annual Meeting of the Association for Computational Linguistics (Volume 1: Long Papers)}, pages 36--53, Bangkok, Thailand. Association for Computational Linguistics.

\bibitem[{Lin and Zhu(2025)}]{Lin25}
Cong~William Lin and Wu~Zhu. 2025.
\newblock \href {https://doi.org/10.48550/ARXIV.2504.13629} {Divergent {LLM} adoption and heterogeneous convergence paths in research writing}.
\newblock \emph{arXiv}, abs/2504.13629.

\bibitem[{Ma et~al.(2023)Ma, Liu, Yi, Cheng, Huang, Lu, and Liu}]{Ma23}
Yongqiang Ma, Jiawei Liu, Fan Yi, Qikai Cheng, Yong Huang, Wei Lu, and Xiaozhong Liu. 2023.
\newblock \href {https://doi.org/10.48550/ARXIV.2301.10416} {{AI} vs. human -- differentiation analysis of scientific content generation}.
\newblock \emph{arXiv}, abs/2301.10416.

\bibitem[{Macko et~al.(2023)Macko, Moro, Uchendu, Lucas, Yamashita, Pikuliak, Srba, Le, Lee, Simko, and Bielikova}]{macko-etal-2023-multitude}
Dominik Macko, Robert Moro, Adaku Uchendu, Jason Lucas, Michiharu Yamashita, Mat{\'u}{\v{s}} Pikuliak, Ivan Srba, Thai Le, Dongwon Lee, Jakub Simko, and Maria Bielikova. 2023.
\newblock \href {https://doi.org/10.18653/v1/2023.emnlp-main.616} {{MULTIT}u{DE}: Large-scale multilingual machine-generated text detection benchmark}.
\newblock In \emph{Proceedings of the 2023 Conference on Empirical Methods in Natural Language Processing}, pages 9960--9987, Singapore. Association for Computational Linguistics.

\bibitem[{Maurer(2026)}]{maurer-2025-elfen}
Maximilian Maurer. 2026.
\newblock elfen: A python package for efficient linguistic feature extraction for natural language datasets.
\newblock In \emph{Proceedings of the 19th Conference of the European Chapter of the Association for Computational Linguistics: System Demonstrations}, Rabat, Morocco. Association for Computational Linguistics.

\bibitem[{Mikolov et~al.(2013)Mikolov, Chen, Corrado, and Dean}]{mikolov2013efficient}
Tom{\'a}s Mikolov, Kai Chen, Greg Corrado, and Jeffrey Dean. 2013.
\newblock \href {http://arxiv.org/abs/1301.3781} {Efficient estimation of word representations in vector space}.
\newblock In \emph{1st {{International Conference}} on {{Learning Representations}}, {{ICLR}} 2013, {{Workshop Track Proceedings}}}, Scottsdale, Arizona, USA.

\bibitem[{Muñoz-Ortiz et~al.(2024)Muñoz-Ortiz, Gómez-Rodríguez, and Vilares}]{MuozOrtiz2024}
Alberto Muñoz-Ortiz, Carlos Gómez-Rodríguez, and David Vilares. 2024.
\newblock \href {https://doi.org/10.1007/s10462-024-10903-2} {Contrasting linguistic patterns in human and {LLM}-generated news text}.
\newblock \emph{Artificial Intelligence Review}, 57(10).

\bibitem[{Pedregosa et~al.(2011)Pedregosa, Varoquaux, Gramfort, Michel, Thirion, Grisel, Blondel, Prettenhofer, Weiss, Dubourg, Vanderplas, Passos, Cournapeau, Brucher, Perrot, and Duchesnay}]{scikit-learn}
Fabian Pedregosa, Ga\"{e}l Varoquaux, Alexandre Gramfort, Vincent Michel, Bertrand Thirion, Olivier Grisel, Mathieu Blondel, Peter Prettenhofer, Ron Weiss, Vincent Dubourg, Jake Vanderplas, Alexandre Passos, David Cournapeau, Matthieu Brucher, Matthieu Perrot, and \'{E}douard Duchesnay. 2011.
\newblock Scikit-learn: {{Machine}} learning in {{Python}}.
\newblock \emph{Journal of Machine Learning Research}, 12:2825--2830.

\bibitem[{Pei et~al.(2022)Pei, Ananthasubramaniam, Wang, Zhou, Dedeloudis, Sargent, and Jurgens}]{pei-etal-2022-potato}
Jiaxin Pei, Aparna Ananthasubramaniam, Xingyao Wang, Naitian Zhou, Apostolos Dedeloudis, Jackson Sargent, and David Jurgens. 2022.
\newblock \href {https://doi.org/10.18653/v1/2022.emnlp-demos.33} {{POTATO}: The portable text annotation tool}.
\newblock In \emph{Proceedings of the 2022 Conference on Empirical Methods in Natural Language Processing: System Demonstrations}, pages 327--337, Abu Dhabi, UAE. Association for Computational Linguistics.

\bibitem[{Pierrejean and Tanguy(2018)}]{pierrejean2018qualitative}
Benedicte Pierrejean and Ludovic Tanguy. 2018.
\newblock \href {https://doi.org/10.18653/v1/N18-4005} {Towards qualitative word embeddings evaluation: {{Measuring}} neighbors variation}.
\newblock In \emph{Proceedings of the 2018 {{Conference}} of the {{North American Chapter}} of the {{Association}} for {{Computational Linguistics}}: {{Student Research}} {{Workshop}}}, pages 32--39, New Orleans, Louisiana, USA. Association for Computational Linguistics.

\bibitem[{Rayson et~al.(2004)Rayson, Berridge, and Francis}]{rayson2004extending}
Paul Rayson, Damon Berridge, and Brian Francis. 2004.
\newblock Extending the cochran rule for the comparison of word frequencies between corpora.
\newblock In \emph{JADT 2004 : 7es Journées internationales d’Analyse statistique des Données Textuelles}.

\bibitem[{Rayson and Garside(2000)}]{rayson-garside-2000-comparing}
Paul Rayson and Roger Garside. 2000.
\newblock \href {https://doi.org/10.3115/1117729.1117730} {Comparing corpora using frequency profiling}.
\newblock In \emph{The Workshop on Comparing Corpora}, pages 1--6, Hong Kong, China. Association for Computational Linguistics.

\bibitem[{{\v R}eh{\r u}{\v r}ek and Sojka(2010)}]{rehurek2010gensim}
Radim {\v R}eh{\r u}{\v r}ek and Petr Sojka. 2010.
\newblock Software framework for topic modelling with large corpora.
\newblock In \emph{Proceedings of the {{LREC}} 2010 Workshop on New Challenges for {{NLP}} Frameworks}, pages 45--50, Valletta, Malta. ELRA.

\bibitem[{Reimers and Gurevych(2019)}]{reimers2019sentencebert}
Nils Reimers and Iryna Gurevych. 2019.
\newblock \href {https://doi.org/10.18653/v1/D19-1410} {Sentence-{{BERT}}: {{Sentence}} embeddings using {{Siamese BERT-Networks}}}.
\newblock In \emph{Proceedings of the 2019 {{Conference}} on {{Empirical Methods}} in {{Natural Language Processing}} and the 9th {{International Joint Conference}} on {{Natural Language Processing}} ({{EMNLP-IJCNLP}})}, pages 3982--3992, Hong Kong, China. Association for Computational Linguistics.

\bibitem[{Reinhart et~al.(2025)Reinhart, Markey, Laudenbach, Pantusen, Yurko, Weinberg, and Brown}]{Reinhart2025}
Alex Reinhart, Ben Markey, Michael Laudenbach, Kachatad Pantusen, Ronald Yurko, Gordon Weinberg, and David~West Brown. 2025.
\newblock \href {https://doi.org/10.1073/pnas.2422455122} {Do {LLMs} write like humans? {Variation} in grammatical and rhetorical styles}.
\newblock \emph{Proceedings of the National Academy of Sciences}, 122(8).

\bibitem[{Richburg et~al.(2024)Richburg, Bao, and Carpuat}]{richburg-etal-2024-automatic}
Aquia Richburg, Calvin Bao, and Marine Carpuat. 2024.
\newblock \href {https://aclanthology.org/2024.lrec-main.165/} {Automatic authorship analysis in human-{AI} collaborative writing}.
\newblock In \emph{Proceedings of the 2024 Joint International Conference on Computational Linguistics, Language Resources and Evaluation (LREC-COLING 2024)}, pages 1845--1855, Torino, Italia. ELRA and ICCL.

\bibitem[{Rohatgi et~al.(2023)Rohatgi, Qin, Aw, Unnithan, and Kan}]{rohatgi-etal-2023-acl}
Shaurya Rohatgi, Yanxia Qin, Benjamin Aw, Niranjana Unnithan, and Min-Yen Kan. 2023.
\newblock \href {https://doi.org/10.18653/v1/2023.emnlp-main.640} {The {ACL} {OCL} corpus: Advancing open science in computational linguistics}.
\newblock In \emph{Proceedings of the 2023 Conference on Empirical Methods in Natural Language Processing}, pages 10348--10361, Singapore. Association for Computational Linguistics.

\bibitem[{Russell et~al.(2025)Russell, Karpinska, and Iyyer}]{russell-etal-2025-people}
Jenna Russell, Marzena Karpinska, and Mohit Iyyer. 2025.
\newblock \href {https://doi.org/10.18653/v1/2025.acl-long.267} {People who frequently use {C}hat{GPT} for writing tasks are accurate and robust detectors of {AI}-generated text}.
\newblock In \emph{Proceedings of the 63rd Annual Meeting of the Association for Computational Linguistics (Volume 1: Long Papers)}, pages 5342--5373, Vienna, Austria. Association for Computational Linguistics.

\bibitem[{Sagi et~al.(2009)Sagi, Kaufmann, and Clark}]{sagi-etal-2009-semantic}
Eyal Sagi, Stefan Kaufmann, and Brady Clark. 2009.
\newblock \href {https://aclanthology.org/W09-0214/} {Semantic density analysis: Comparing word meaning across time and phonetic space}.
\newblock In \emph{Proceedings of the Workshop on Geometrical Models of Natural Language Semantics}, pages 104--111, Athens, Greece. Association for Computational Linguistics.

\bibitem[{Schlechtweg(2023)}]{Schlechtweg2023measurement}
Dominik Schlechtweg. 2023.
\newblock \href {http://dx.doi.org/10.18419/opus-12833} {\emph{Human and Computational Measurement of Lexical Semantic Change}}.
\newblock Stuttgart, Germany.

\bibitem[{Su et~al.(2025)Su, Wang, Wan, Zhang, and Luo}]{su-etal-2025-haco}
Zhixiong Su, Yichen Wang, Herun Wan, Zhaohan Zhang, and Minnan Luo. 2025.
\newblock \href {https://doi.org/10.18653/v1/2025.acl-long.1069} {{HAC}o-det: A study towards fine-grained machine-generated text detection under human-{AI} coauthoring}.
\newblock In \emph{Proceedings of the 63rd Annual Meeting of the Association for Computational Linguistics (Volume 1: Long Papers)}, pages 22015--22036, Vienna, Austria. Association for Computational Linguistics.

\bibitem[{Tahmasebi et~al.(2021)Tahmasebi, Borin, and Jatowt}]{tahmasebi2021survey}
Nina Tahmasebi, Lars Borin, and Adam Jatowt. 2021.
\newblock \href {https://doi.org/10.5281/zenodo.5040302} {Survey of computational approaches to lexical semantic change}.
\newblock In Nina Tahmasebi, Lars Borin, Adam Jatowt, Yang Xu, and Simon Hengchen, editors, \emph{Computational {{Approaches}} to {{Semantic Change}}}, pages 1--91. Language Science Press, Berlin.

\bibitem[{Wang et~al.(2024)Wang, Mansurov, Ivanov, Su, Shelmanov, Tsvigun, Whitehouse, Mohammed~Afzal, Mahmoud, Sasaki, Arnold, Aji, Habash, Gurevych, and Nakov}]{wang-etal-2024-m4}
Yuxia Wang, Jonibek Mansurov, Petar Ivanov, Jinyan Su, Artem Shelmanov, Akim Tsvigun, Chenxi Whitehouse, Osama Mohammed~Afzal, Tarek Mahmoud, Toru Sasaki, Thomas Arnold, Alham~Fikri Aji, Nizar Habash, Iryna Gurevych, and Preslav Nakov. 2024.
\newblock \href {https://doi.org/10.18653/v1/2024.eacl-long.83} {M4: Multi-generator, multi-domain, and multi-lingual black-box machine-generated text detection}.
\newblock In \emph{Proceedings of the 18th Conference of the European Chapter of the Association for Computational Linguistics (Volume 1: Long Papers)}, pages 1369--1407, St. Julian{'}s, Malta. Association for Computational Linguistics.

\bibitem[{Wang et~al.(2025)Wang, Shelmanov, Mansurov, Tsvigun, Habash, Aji, Artemova, Xie, Su, Xing, Gurevych, and Nakov}]{coauthorbenchmark}
Yuxia Wang, Artem Shelmanov, Jonibek Mansurov, Akim Tsvigun, Nizar Habash, Alham~Fikri Aji, Ekaterina Artemova, Zhuohan Xie, Jinyan Su, Rui Xing, Iryna Gurevych, and Preslav Nakov. 2025.
\newblock \href {https://doi.org/10.5281/ZENODO.14966981} {{PAN'25} generative {AI} detection (task 2): Human-{AI} collaborative text classification}.

\bibitem[{Warner et~al.(2025)Warner, Chaffin, Clavi{\'e}, Weller, Hallstr{\"o}m, Taghadouini, Gallagher, Biswas, Ladhak, Aarsen, Adams, Howard, and Poli}]{modernbert}
Benjamin Warner, Antoine Chaffin, Benjamin Clavi{\'e}, Orion Weller, Oskar Hallstr{\"o}m, Said Taghadouini, Alexis Gallagher, Raja Biswas, Faisal Ladhak, Tom Aarsen, Griffin~Thomas Adams, Jeremy Howard, and Iacopo Poli. 2025.
\newblock \href {https://doi.org/10.18653/v1/2025.acl-long.127} {Smarter, better, faster, longer: A modern bidirectional encoder for fast, memory efficient, and long context finetuning and inference}.
\newblock In \emph{Proceedings of the 63rd Annual Meeting of the Association for Computational Linguistics (Volume 1: Long Papers)}, pages 2526--2547, Vienna, Austria. Association for Computational Linguistics.

\bibitem[{Wu et~al.(2025)Wu, Yang, Zhan, Yuan, Chao, and Wong}]{wu-etal-2025-survey}
Junchao Wu, Shu Yang, Runzhe Zhan, Yulin Yuan, Lidia~Sam Chao, and Derek~Fai Wong. 2025.
\newblock \href {https://doi.org/10.1162/coli_a_00549} {A survey on {LLM}-generated text detection: Necessity, methods, and future directions}.
\newblock \emph{Computational Linguistics}, 51(1):275--338.

\bibitem[{Yildiz~Durak et~al.(2025)Yildiz~Durak, Eğin, and Onan}]{YildizDurak2025}
Hatice Yildiz~Durak, Figen Eğin, and Aytuğ Onan. 2025.
\newblock \href {https://doi.org/10.1111/ejed.70014} {A comparison of human‐written versus {AI}‐generated text in discussions at educational settings: {Investigating} features for {ChatGPT}, {Gemini} and {BingAI}}.
\newblock \emph{European Journal of Education}, 60(1).

\bibitem[{Zamaraeva et~al.(2025)Zamaraeva, Flickinger, Bond, and G{\'o}mez-Rodr{\'i}guez}]{zamaraeva-etal-2025-comparing}
Olga Zamaraeva, Dan Flickinger, Francis Bond, and Carlos G{\'o}mez-Rodr{\'i}guez. 2025.
\newblock \href {https://doi.org/10.18653/v1/2025.acl-long.443} {Comparing {LLM}-generated and human-authored news text using formal syntactic theory}.
\newblock In \emph{Proceedings of the 63rd Annual Meeting of the Association for Computational Linguistics (Volume 1: Long Papers)}, pages 9041--9060, Vienna, Austria. Association for Computational Linguistics.

\bibitem[{Zanotto and Aroyehun(2024)}]{zanotto24}
Sergio~E. Zanotto and Segun Aroyehun. 2024.
\newblock \href {https://doi.org/10.48550/ARXIV.2412.03025} {Human variability vs. machine consistency: A linguistic analysis of texts generated by humans and large language models}.
\newblock \emph{arXiv}, abs/2412.03025.

\bibitem[{Zhao et~al.(2025)Zhao, Gunn, Christ, Fairoze, Fabrega, Carlini, Garg, Hong, Nasr, Tram{\`{e}}r, Jha, Li, Wang, and Song}]{watermarking}
Xuandong Zhao, Sam Gunn, Miranda Christ, Jaiden Fairoze, Andres Fabrega, Nicholas Carlini, Sanjam Garg, Sanghyun Hong, Milad Nasr, Florian Tram{\`{e}}r, Somesh Jha, Lei Li, Yu{-}Xiang Wang, and Dawn Song. 2025.
\newblock \href {https://doi.org/10.1109/SP61157.2025.00178} {{SoK}: Watermarking for {AI}-generated content}.
\newblock In \emph{{IEEE} Symposium on Security and Privacy, {SP} 2025, San Francisco, CA, USA, May 12-15, 2025}, pages 2621--2639. {IEEE}.

\end{thebibliography}

% \section{Language Resource References}
% \label{lr:ref}
% \bibliographystylelanguageresource{lrec2026-natbib}
% \bibliographylanguageresource{languageresource}

\clearpage
\appendix
\section{Topic analysis}
\label{app:topic}

We inspect the comparability of the two time periods by analyzing their distribution across topics.
% The initial corpus includes topic labels from a classifier created by the original authors,
% but we were unfortunately unable to locate the model or the experimental settings to reproduce it.
% We therefore implement a new pipeline based on the more recent BERTopic model \citep{grootendorst2022bertopic}.
We rely on BERTopic \citep{grootendorst2022bertopic}:
given a set of documents, it computes document embeddings using a pretrained transformer model, clusters those embeddings, and then represents the topics (corresponding to the obtained clusters) by identifying a set of distinctive keywords.
We set the minimum topic size to 100 documents, representation model to Maximal Marginal Relevance, and use default values for other parameters.
%We represent each paper as the concatenation of its title and abstract,
For each paper, we use the concatenation of its title and abstract.
%and fit the topic model to the papers from our target time periods (as defined above). %(as defined in Section~\ref{sec:corpus-structure}).
Papers initially classified as outliers are assigned to the best-fitting topic based on the probabilities computed in the soft-clustering step over document embeddings.
We include the trained topic model in our corpus update pipeline 
to ensure consistency of topic labels for future papers.

% \paragraph{Text preprocessing} The original corpus only provides the text extracted by GROBID without further preprocessing.
For a given topic, we calculate what proportion of papers assigned to it come from \tone vs.\ \ttwo (after normalizing the counts by the total number of papers in the respective time period).
We show sample topics with different temporal distributions in Table~\ref{tab:topic-distribution}.
Topics which are overrepresented in one of the two periods reflect general shifts in research trends within NLP.
For example, papers focusing on individual levels of linguistic structure (e.g., word sense disambiguation, topic 41; dependency parsing, 45) and related methods (word embeddings, 19) are more frequent in \tone.
Those concerned with LLMs (13, 54) and more recent methods such as reinforcement learning (62) are more prevalent in \ttwo.
But beyond these rather intuitive differences, 39 out of 70 topics (62\% of all papers) have a broadly balanced temporal distribution (normalized proportion of papers from the dominant time period $\leq$\,60\%).
Together with the fact that strictly all topics contain papers from both \tone and \ttwo, we interpret these findings as confirming the overall comparability of the two time periods, without stark topical shifts likely to skew the outcome of our experiments.

\begin{table}[t]
    \centering
    \resizebox{\linewidth}{!}{%
    \begin{tabular}{lrrr}
    \toprule
    \textbf{Topic} & \textbf{\% t\textsubscript{1}} & \textbf{\% t\textsubscript{2}} & \textbf{Total} \\
    \midrule
    67\_humor\_humorous\_offense\_pun & 79.8 & 20.2 & 117 \\
    51\_sense\_word\_wsd\_disambiguation & 68.1 & 31.9 & 333 \\
    45\_treebank\_ud\_universal\_dependency & 66.3 & 33.7 & 495 \\
    19\_embeddings\_similarity\_word\_sentence & 65.5 & 34.5 & 637 \\
    \midrule
    61\_annotation\_nlp\_annotators\_and & 52.8 & 47.2 & 342 \\
    15\_question\_questions\_qa\_answering & 50.1 & 49.9 & 1,126 \\
    11\_crosslingual\_languages\_multilingual\_language & 49.7 & 50.3 & 962 \\
    36\_claim\_claims\_evidence\_verification & 49.3 & 50.7 & 236 \\
    \midrule
    43\_text\_machinegenerated\_authorship\_detection & 26.4 & 73.6 & 208 \\
    13\_reasoning\_llms\_mathematical\_cot & 24.3 & 75.7 & 792 \\
    54\_hallucination\_hallucinations\_llms\_detection & 15.0 & 85.0 & 171 \\
    62\_reward\_preference\_alignment\_rlhf & 4.6 & 95.4 & 152 \\
    \bottomrule
    \end{tabular}}
    \caption{Sample topics with different temporal distribution. Percentages show the normalized proportion of papers from each time period within a topic; total shows topic size as the raw number of papers.}
    \label{tab:topic-distribution}
\end{table}

% TOTAL: 70 topics, 37914 papers

% T1-DOM
% topics: 15 (21.4)
% papers: 5033 (13.3)

% MIDDLE
% topics: 39 (55.7)
% papers: 23542 (62.1)

% T2-DOM
% topics: 16 (22.9)
% papers: 9339 (24.6)

\section{Additional Linguistic Examples}
\label{app:lexical}
We present additional examples from our lexical analysis introduced in Section~\ref{sec:lexical-choices}.
We provide more detailed corpus statistics for single-word lexical choices; 
extend the same analysis to multi-word sequences, operationalized as word 5-grams;
and then elaborate on our clustering analysis by discussing more target words and sample sentences.

\subsection{Single-Word Lexical Choices}
We present the strongest changes in single-word lexical choices for nouns (Table~\ref{tab:app-nouns}), adjectives (Table~\ref{tab:app-adjectives}), verbs (Table~\ref{tab:app-verbs}), and adverbs (Table~\ref{tab:app-adverbs}).
Each table shows the 10 words with the strongest increase (top panel) and decrease (bottom panel) in frequency over time, as measured by the log-likelihood score.
Examples are shown both for the naturalistic corpus (left panel) and the LLM-modified corpus (right panel).
The following columns are shown:
\begin{itemize}[nosep]
\item $Freq_{t_1}$ -- frequency per million words in \tone 
\item $Freq_{t_2}$ -- frequency per million words in \ttwo
\item $LL$ -- log-likelihood score
\item $ND_{t_1}$ -- neighborhood density in \tone
\item $ND_{t_2}$ -- neighborhood density in \ttwo
\item $\Delta ND$ -- change in neighborhood density in \ttwo
\item $U$ -- Mann-Whitney--U test statistic for comparison of neighborhood densities
\item $p$ -- significance levels for the Mann-Whitney--U test: ***$<0.001$; **$<0.01$; *$<0.05$; ns\,$\ge0.05$
\end{itemize}

% The following columns are shown.
% $Freq_{t_1}$: frequency per million words in \tone; 
% $Freq_{t_2}$: frequency per million words in \ttwo;
% $LL$: log-likelihood score; 
% $ND_{t_1}$: neighborhood density in \tone;
% $ND_{t_2}$: neighborhood density in \ttwo;
% $\Delta ND$: change in neighborhood density in \ttwo;
% $U$: Mann-Whitney--U test statistic for comparison of neighborhood densities;
% $p$: p-value for the test expressed as levels of significance: ***$<0.001$; **$<0.01$; *$<0.05$; ns\,$\ge0.05$.
% Columns: frequency per million words in \tone (Freq\,$_{t_1}$) and \ttwo (Freq\,$_{t_2}$); log-likelihood score (LL); neighborhood density in \tone ($ND_{t_1}$), \ttwo ($ND_{t_2}$), and the difference between the two ($\Delta ND$); Mann-Whitney--U test statistic (U) and p-value (p) for comparison of neighborhood densities (expressed as levels of significance: ***$<0.001$; **$<0.01$; *$<0.05$; ns\,$\ge0.05$).
Naturalistic corpus targets marked with an asterisk also appear in the top 100 strongest changes in the LLM-modified corpus (for the respective part of speech and increase/decrease direction).

\subsection{Multi-Word Lexical Choices}
We perform a follow-up to our core analysis of single-word lexical choices, aiming to understand if comparable patterns can also be observed in longer sequences of words.
We operationally define these as word 5-grams (lemmatized and part-of-speech tagged),
retaining only those where each of the five constituent lemmas is composed of alphabetic characters only and is at least two characters long.
We then collect frequency counts in \tone and \ttwo for both the naturalistic and the LLM-modified corpus, and compute the log-likelihood score using the same procedure as for individual words.
We restrict our analysis to the 5-grams which appear at least 10 times in both \tone and \ttwo (for the corresponding corpus).
The results are summarized in Table~\ref{tab:app-fivegrams}, which shows the 40 strongest rises and falls in use for the naturalistic and the LLM-modified corpus.

Some of the word sequences identified by the analysis point to the general evolution of the NLP community in terms of dominant methods (e.g., \textit{language model such as bert} falling out of use) and writing conventions (e.g., \textit{name or uniquely identify individual} growing in use, presumably as part of the Responsible NLP Checklist).
But many other sequences clearly reflect the stylistic patterns already observed for single words.
For example, the naturalistic corpus shows a decrease in meta-narrative devices relying on simple vocabulary (e.g., \textit{we can see that the}, \textit{result show that our model}) and an increase in more formal equivalent expressions (e.g., \textit{it be evident that the}, \textit{provide valuable insight into}).
The LLM-modified corpus shows the same trend, with straightforward expressions falling out of use (e.g., \textit{paper be organize as follow}, \textit{it have be show that}), and their formal equivalents becoming more prominent (e.g., \textit{paper be structure as follow}, \textit{it have be observe that}).
Like in the single-word analysis, we overall see that the naturalistic and the LLM-modified corpus overlap in most prominent stylistic changes,
and that these generally involve more complex lexical choices.
These findings further support the view that the stylistic changes in the naturalistic corpus can be at least partly attributed to LLM-assisted writing.

\subsection{Clustering Examples}
We provide further examples from our clustering analysis to illustrate fine-grained usage differences for the following target words:
\textit{ensure} (Table~\ref{tab:clust-ensure}),
\textit{utilize} (Table~\ref{tab:clust-utilize}),
\textit{crucial} (Table~\ref{tab:clust-crucial}),
and \textit{notably} (Table~\ref{tab:clust-notably}).
For each target word, we include two sample clusters capturing distinct uses, with four representative sentences manually selected for each cluster.
For ease of reading, the examples are shown in a keyword-in-context format.

\section{Human Annotation}\label{app:annotation}
\begin{figure*}
    \centering
    \includegraphics[width=0.99\linewidth]{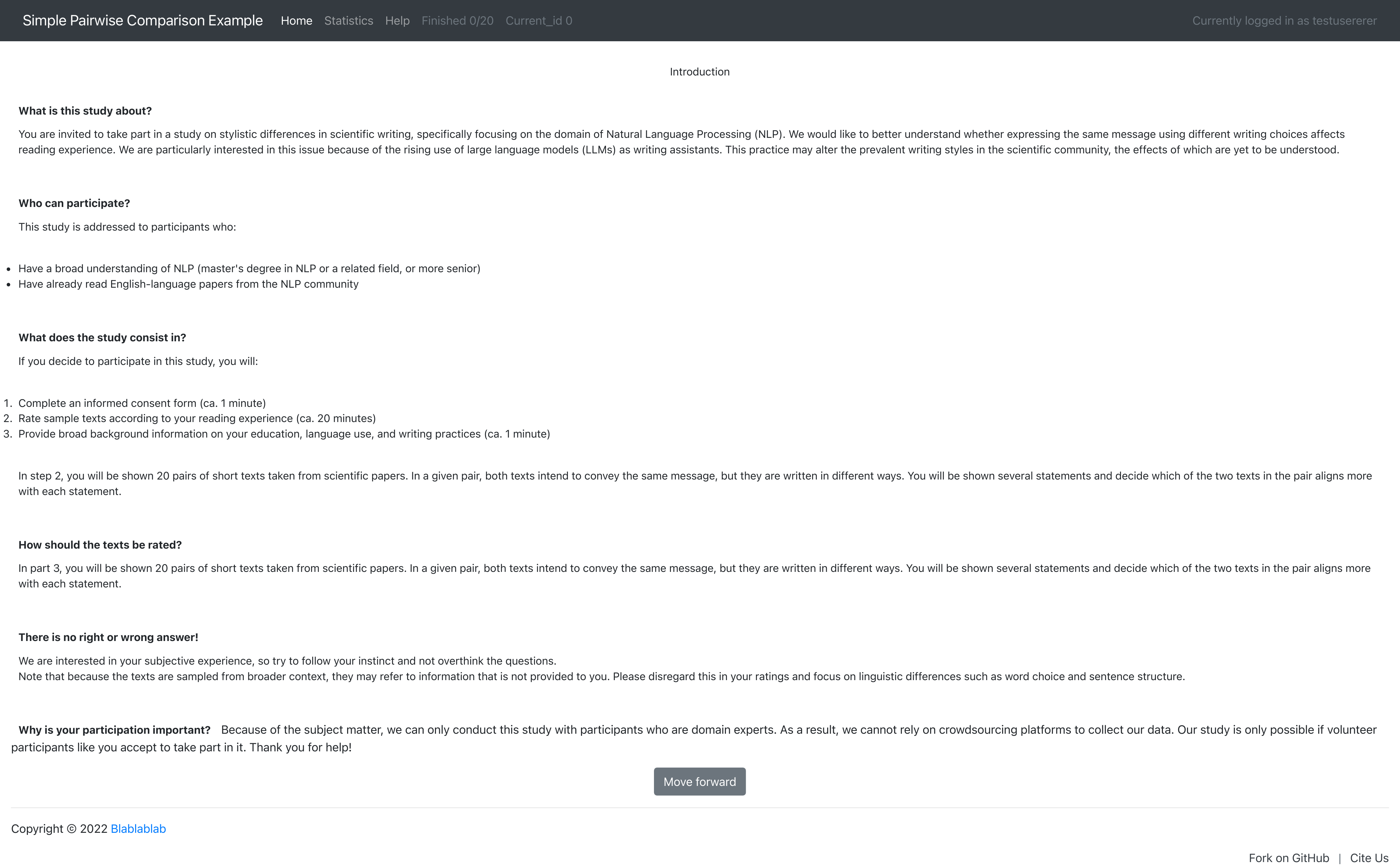}
    \caption{The introduction and study description annotators saw before rating the pairs of human vs.\ LLM-paraphrased scientific text.}
    \label{fig:studyintro}
\end{figure*}

\begin{figure*}
    \centering
    \includegraphics[width=0.99\linewidth]{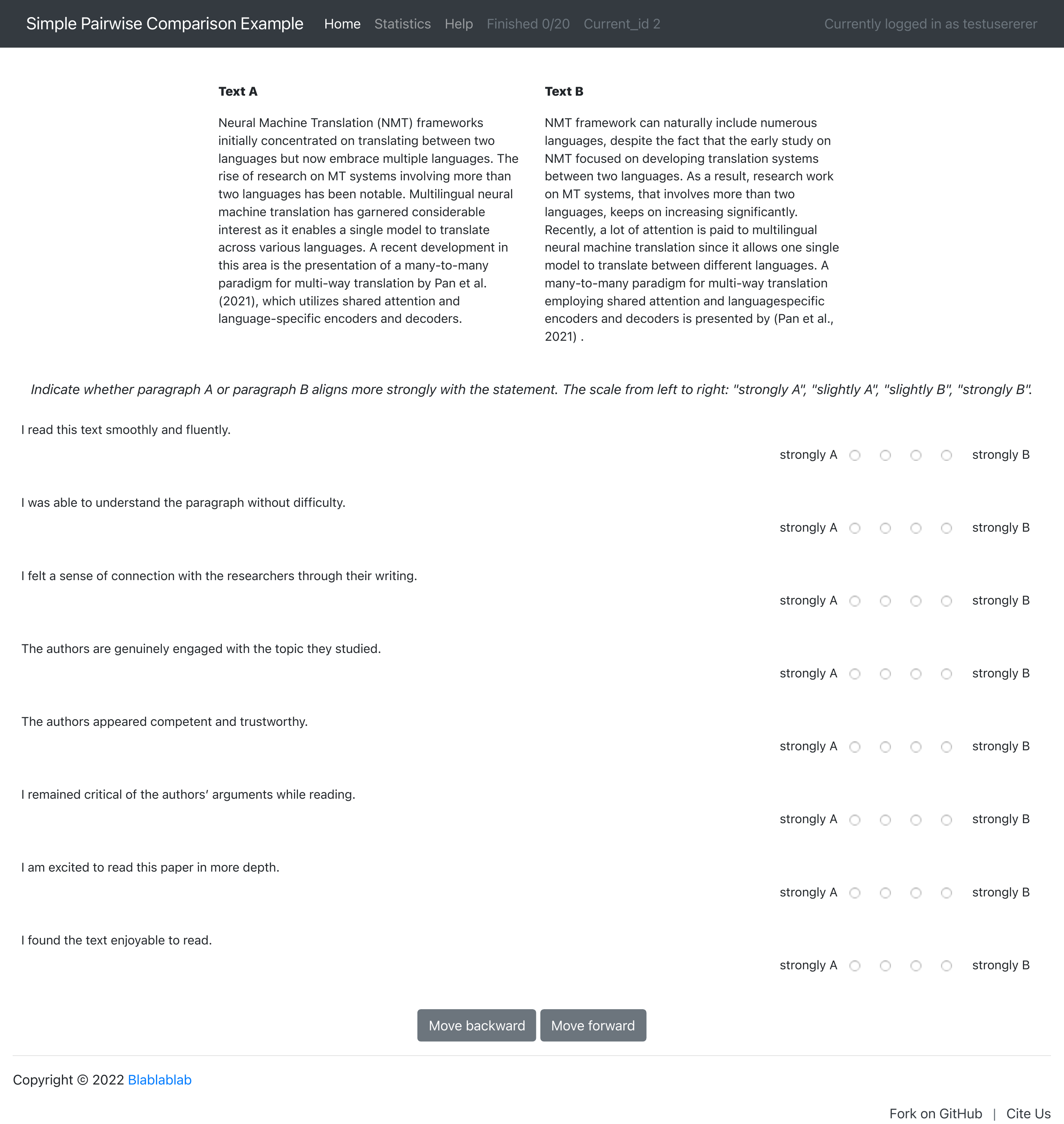}
    \caption{View of an example annotation item: pairwise annotation on four different dimensions of reading experience.}
    \label{fig:studyview}
\end{figure*}

In the human annotation, we measured four different dimensions of reading experience, each recorded with 2 items. The items for all dimensions are listed in Table~\ref{tab:readingexperience}. The annotation guidelines are presented in Figure~\ref{fig:studyintro} and an example of one item and the annotator view is shown in Figure~\ref{fig:studyview}. 
Annotation was conducted using the Potato annotation tool \citep{pei-etal-2022-potato}.
The full configuration, dataset and annotation results are provided via the repository: \url{https://github.com/FilipMiletic/ScientificCommunication }

\begin{table*}
    \centering
    \resizebox{\textwidth}{!}{%
    \begin{tabular}{llrrrrrrrl|lrrr}
    \toprule
    \multicolumn{10}{c|}{\textbf{Naturalistic corpus}} & \multicolumn{4}{c}{\textbf{LLM-modified corpus}} \\
    \midrule
     & Target & $Freq_{t_1}$ & $Freq_{t_2}$ & \multicolumn{1}{c}{$LL$} & $ND_{t_1}$ & $ND_{t_2}$ & $\Delta ND$ &   \multicolumn{1}{c}{$U$} & \multicolumn{1}{c|}{$p$} & Target &  $Freq_{t_1}$ & $Freq_{t_2}$ & \multicolumn{1}{c}{$LL$} \\
        \midrule
        &         prompt &      169.2 &      910.0 & 56679.7 & 0.620 & 0.626 &    0.006 & 4122.0 &    * &         study &      879.5 &     2249.8 & 303.0 \\
        &            llm &        1.9 &      403.9 & 54465.4 & 0.618 & 0.653 &    0.036 & 1834.0 &  *** &      approach &     1389.7 &     2565.0 & 173.0 \\
        &    instruction &       94.9 &      613.6 & 43024.1 & 0.567 & 0.575 &    0.008 & 4038.0 &    * &     challenge &      397.9 &     1079.6 & 159.5 \\
        &      reasoning &      223.9 &      593.3 & 17605.8 & 0.632 & 0.675 &    0.043 & 2455.5 &  *** &      research &      757.1 &     1514.3 & 125.6 \\
      {*} &     capability &       83.3 &      339.2 & 16869.6 & 0.629 & 0.647 &    0.019 & 3248.0 &  *** &   advancement &       12.2 &      216.3 & 108.1 \\
        &           shot &      251.4 &      559.4 & 12201.6 & 0.661 & 0.689 &    0.028 & 3314.5 &  *** &         realm &       12.2 &      171.0 &  80.2 \\
      {*} &     limitation &      138.3 &      360.5 & 10416.0 & 0.601 & 0.586 &   -0.015 & 6022.5 &    * &       process &      518.3 &      991.0 &  73.5 \\
      {*} &       response &      391.5 &      722.1 & 10131.6 & 0.628 & 0.589 &   -0.039 & 8138.0 &  *** &   utilization &       12.2 &      156.6 &  71.4 \\
        &            cot &        0.1 &       71.5 &  9866.6 & 0.756 & 0.726 &   -0.029 & 7505.0 &  *** &      instance &      546.9 &      956.0 &  55.0 \\
        &  demonstration &       23.5 &      143.5 &  9714.8 & 0.571 & 0.637 &    0.066 &  785.5 &  *** &   methodology &      124.5 &      344.1 &  52.2 \\
\midrule
        &           word &     2879.9 &     1485.3 & 46279.9 & 0.630 & 0.647 &    0.017 & 3028.0 &  *** &          work &     2036.5 &      976.6 & 185.7 \\
        &       sentence &     2502.5 &     1619.7 & 19432.9 & 0.600 & 0.618 &    0.018 & 3073.5 &  *** &       problem &      765.2 &      288.4 & 109.1 \\
        &      embedding &      812.4 &      416.1 & 13283.1 & 0.661 & 0.664 &    0.002 & 4901.5 &   ns &           way &      477.5 &      191.6 &  61.5 \\
        &        network &      526.5 &      255.8 &  9758.4 & 0.677 & 0.676 &   -0.001 & 5185.0 &   ns &          kind &      106.1 &       12.4 &  41.4 \\
        & representation &     1190.7 &      764.5 &  9550.2 & 0.617 & 0.654 &    0.037 & 2025.0 &  *** &      addition &      297.9 &      138.0 &  29.3 \\
        &         vector &      577.0 &      297.3 &  9292.2 & 0.685 & 0.694 &    0.009 & 4240.5 &   ns &       setting &      457.1 &      265.8 &  25.0 \\
        &         corpus &      706.2 &      398.1 &  8887.6 & 0.645 & 0.640 &   -0.005 & 5992.5 &    * &          idea &      167.3 &       61.8 &  24.6 \\
        &      attention &      767.4 &      447.1 &  8721.4 & 0.644 & 0.639 &   -0.005 & 5114.0 &   ns &       respect &       95.9 &       24.7 &  21.9 \\
        &          layer &      942.3 &      608.9 &  7366.7 & 0.695 & 0.703 &    0.008 & 4291.5 &   ns &     intuition &       38.8 &        2.1 &  19.6 \\
        &        feature &     1043.3 &      696.8 &  7084.2 & 0.603 & 0.636 &    0.034 & 1606.5 &  *** &        amount &      279.6 &      152.5 &  18.5 \\
\bottomrule
\end{tabular}}
    \caption{Strongest changes in \textbf{noun} usage.}
    \label{tab:app-nouns}
\end{table*}

\begin{table*}
    \centering
    \resizebox{\textwidth}{!}{%
    \begin{tabular}{llrrrrrrrl|lrrr}
    \toprule
    \multicolumn{10}{c|}{\textbf{Naturalistic corpus}} & \multicolumn{4}{c}{\textbf{LLM-modified corpus}} \\
    \midrule
     & Target & $Freq_{t_1}$ & $Freq_{t_2}$ & \multicolumn{1}{c}{$LL$} & $ND_{t_1}$ & $ND_{t_2}$ & $\Delta ND$ &   \multicolumn{1}{c}{$U$} & \multicolumn{1}{c|}{$p$} & Target &  $Freq_{t_1}$ & $Freq_{t_2}$ & \multicolumn{1}{c}{$LL$} \\
\midrule
        &         prompt &       55.9 &      268.0 & 15324.2 & 0.603 & 0.608 &    0.005 & 3950.0 &    * &       various &      461.2 &     1473.1 & 271.7 \\
        &            llm &        0.6 &       61.4 &  8023.0 & 0.617 & 0.605 &   -0.012 & 7262.5 &  *** &   significant &      212.2 &      684.0 & 127.6 \\
      {*} &  comprehensive &       51.5 &      175.7 &  7298.5 & 0.606 & 0.634 &    0.028 & 2586.0 &  *** &      distinct &       65.3 &      304.9 &  82.2 \\
      {*} &        diverse &      130.0 &      255.5 &  4232.5 & 0.584 & 0.598 &    0.014 & 3564.0 &  *** & comprehensive &      104.1 &      379.1 &  81.2 \\
      {*} &        various &      304.9 &      483.4 &  4148.1 & 0.602 & 0.600 &   -0.002 & 5207.0 &   ns &       primary &       55.1 &      276.1 &  78.6 \\
      {*} &    significant &      258.7 &      414.8 &  3710.9 & 0.590 & 0.587 &   -0.003 & 5050.5 &   ns &       crucial &      112.2 &      387.3 &  78.2 \\
        &          blank &       11.2 &       57.6 &  3475.3 & 0.581 & 0.530 &   -0.051 & 8354.5 &  *** &      specific &      834.6 &     1403.0 &  71.2 \\
        &    demographic &       28.8 &       86.8 &  3094.5 & 0.642 & 0.613 &   -0.028 & 7095.5 &  *** &      valuable &       42.9 &      206.0 &  56.8 \\
      {*} &      potential &      108.8 &      203.2 &  2949.4 & 0.606 & 0.615 &    0.009 & 4021.5 &    * &      superior &       16.3 &      140.1 &  54.8 \\
      {*} &        crucial &       70.0 &      147.4 &  2869.0 & 0.618 & 0.600 &   -0.018 & 6768.5 &  *** &       notable &       34.7 &      179.2 &  52.2 \\
\midrule
        &         neural &      427.0 &      175.5 & 11057.2 & 0.666 & 0.657 &   -0.010 & 5946.0 &    * &          well &      363.2 &      109.2 &  70.2 \\
        &      syntactic &      265.0 &      127.0 &  5068.2 & 0.691 & 0.700 &    0.010 & 4106.5 &    * &         above &      146.9 &       14.4 &  61.6 \\
        &           lstm &       60.8 &       12.6 &  3525.7 & 0.737 & 0.690 &   -0.047 & 8028.5 &  *** &     different &     1669.2 &     1083.7 &  61.2 \\
        &           bert &      146.8 &       63.8 &  3434.0 & 0.753 & 0.762 &    0.009 & 3733.5 &   ** &          good &      542.8 &      241.0 &  58.1 \\
        &   unsupervised &      167.7 &       80.3 &  3213.0 & 0.639 & 0.641 &    0.002 & 4489.0 &   ns &          able &      208.1 &       43.3 &  57.3 \\
        &        lexical &      260.6 &      155.3 &  2744.9 & 0.642 & 0.639 &   -0.003 & 5406.5 &   ns &          many &      673.4 &      362.6 &  46.2 \\
        &       semantic &      613.7 &      461.9 &  2192.3 & 0.648 & 0.667 &    0.019 & 3263.5 &  *** &          hard &      175.5 &       47.4 &  38.2 \\
        &  morphological &      107.9 &       50.5 &  2171.7 & 0.694 & 0.708 &    0.014 & 3817.0 &   ** &        useful &      167.3 &       45.3 &  36.3 \\
      {*} &           same &      966.0 &      783.6 &  1943.4 & 0.507 & 0.520 &    0.013 & 2925.5 &  *** &      possible &      351.0 &      158.6 &  36.3 \\
      {*} &           good &      743.9 &      587.2 &  1885.7 & 0.563 & 0.546 &   -0.018 & 6990.0 &  *** &    particular &      291.8 &      127.7 &  32.1 \\
\bottomrule
\end{tabular}}
    \caption{Strongest changes in \textbf{adjective} usage.}
    \label{tab:app-adjectives}
\end{table*}

\begin{table*}
    \centering
    \resizebox{\textwidth}{!}{%
    \begin{tabular}{llrrrrrrrl|lrrr}
    \toprule
    \multicolumn{10}{c|}{\textbf{Naturalistic corpus}} & \multicolumn{4}{c}{\textbf{LLM-modified corpus}} \\
    \midrule
     & Target & $Freq_{t_1}$ & $Freq_{t_2}$ & \multicolumn{1}{c}{$LL$} & $ND_{t_1}$ & $ND_{t_2}$ & $\Delta ND$ &   \multicolumn{1}{c}{$U$} & \multicolumn{1}{c|}{$p$} & Target &  $Freq_{t_1}$ & $Freq_{t_2}$ & \multicolumn{1}{c}{$LL$} \\
 \midrule
        &         prompt &       33.1 &      303.0 & 25305.3 & 0.612 & 0.665 &    0.053 & 1399.5 &  *** &       utilize &      257.1 &     1905.7 & 693.9 \\
      {*} &        enhance &      126.9 &      403.2 & 15384.0 & 0.643 & 0.636 &   -0.007 & 6045.5 &    * &       enhance &      197.9 &     1306.2 & 446.2 \\
        &       generate &     1248.9 &     1752.5 &  8640.3 & 0.605 & 0.598 &   -0.007 & 5817.5 &    * &       involve &      240.8 &     1291.8 & 386.6 \\
      {*} &        utilize &      222.1 &      454.0 &  8254.0 & 0.631 & 0.605 &   -0.026 & 7447.5 &  *** &       address &      338.7 &     1112.5 & 212.2 \\
      {*} &        exhibit &       63.3 &      202.8 &  7813.6 & 0.568 & 0.569 &    0.001 & 5094.0 &   ns &     introduce &      691.8 &     1633.8 & 191.7 \\
      {*} &         employ &      218.3 &      435.5 &  7476.5 & 0.586 & 0.573 &   -0.014 & 6577.5 &  *** &        assess &      122.4 &      587.2 & 161.6 \\
      {*} &         ensure &      139.9 &      321.5 &  7460.6 & 0.536 & 0.549 &    0.013 & 3791.5 &   ** &        employ &      222.4 &      782.9 & 161.6 \\
      {*} &         assess &      119.0 &      276.9 &  6582.3 & 0.617 & 0.614 &   -0.003 & 5176.5 &   ns &   demonstrate &      351.0 &      953.9 & 141.3 \\
      {*} &    demonstrate &      313.2 &      540.9 &  6250.3 & 0.624 & 0.616 &   -0.008 & 6262.0 &   ** &     encompass &       14.3 &      276.1 & 141.0 \\
        &     underscore &        2.5 &       53.5 &  5819.9 & 0.598 & 0.598 &    0.001 & 5082.5 &   ns &   incorporate &      242.8 &      729.3 & 124.4 \\
\midrule
      {*} &          learn &     1037.9 &      719.1 &  5934.0 & 0.581 & 0.583 &    0.002 & 4519.0 &   ns &          show &     1552.9 &      661.3 & 180.0 \\
        &          train &     1838.0 &     1415.0 &  5624.7 & 0.610 & 0.631 &    0.021 & 2670.5 &  *** &           use &     4395.5 &     2933.8 & 143.1 \\
        &          embed &      603.9 &      380.0 &  5245.9 & 0.662 & 0.673 &    0.010 & 4053.0 &    * &          have &     1283.6 &      665.5 &  97.2 \\
      {*} &            use &     5279.8 &     4674.5 &  3754.2 & 0.545 & 0.481 &   -0.064 & 8634.5 &  *** &          give &     1169.3 &      593.4 &  93.4 \\
        &         encode &      253.5 &      151.4 &  2654.2 & 0.627 & 0.648 &    0.021 & 2854.5 &  *** &       propose &     1763.1 &     1100.2 &  75.5 \\
        &        predict &      682.2 &      517.2 &  2322.6 & 0.570 & 0.578 &    0.009 & 3760.0 &   ** &       perform &      640.8 &      272.0 &  74.7 \\
        &           hide &      147.7 &       79.1 &  2149.9 & 0.661 & 0.656 &   -0.005 & 5378.0 &   ns &        obtain &      518.3 &      206.0 &  67.8 \\
      {*} &          model &      232.8 &      146.6 &  2014.8 & 0.612 & 0.613 &    0.001 & 4611.0 &   ns &           get &      167.3 &       18.5 &  66.7 \\
      {*} &       describe &      519.1 &      386.4 &  1989.3 & 0.610 & 0.609 &   -0.001 & 4886.5 &   ns &        ignore &      120.4 &        8.2 &  57.0 \\
      {*} &           have &     1624.6 &     1391.3 &  1840.6 & 0.505 & 0.489 &   -0.016 & 6814.0 &  *** &      finetune &      120.4 &       12.4 &  49.6 \\
\bottomrule
\end{tabular}}
    \caption{Strongest changes in \textbf{verb} usage.}
    \label{tab:app-verbs}
\end{table*}

\begin{table*}
    \centering
    \resizebox{\textwidth}{!}{%
    \begin{tabular}{llrrrrrrrl|lrrr}
    \toprule
    \multicolumn{10}{c|}{\textbf{Naturalistic corpus}} & \multicolumn{4}{c}{\textbf{LLM-modified corpus}} \\
    \midrule
     & Target & $Freq_{t_1}$ & $Freq_{t_2}$ & \multicolumn{1}{c}{$LL$} & $ND_{t_1}$ & $ND_{t_2}$ & $\Delta ND$ &   \multicolumn{1}{c}{$U$} & \multicolumn{1}{c|}{$p$} & Target &  $Freq_{t_1}$ & $Freq_{t_2}$ & \multicolumn{1}{c}{$LL$} \\
 \midrule
      {*} &   additionally &      137.2 &      281.0 &  5136.8 & 0.605 & 0.620 &    0.015 & 4576.5 &   ns &  additionally &      140.8 &      815.9 & 257.3 \\
      {*} &    effectively &       98.9 &      209.3 &  4114.3 & 0.656 & 0.637 &   -0.019 & 7286.0 &  *** &     initially &       12.2 &      255.5 & 132.7 \\
      {*} &        notably &       33.8 &      100.6 &  3524.0 & 0.600 & 0.629 &    0.029 & 3499.5 &  *** &   effectively &      126.5 &      488.3 & 110.9 \\
      {*} &        thereby &       30.0 &       90.3 &  3206.5 & 0.564 & 0.561 &   -0.003 & 5295.5 &   ns &       notably &       36.7 &      269.9 &  97.8 \\
      {*} &   subsequently &       27.6 &       82.5 &  2915.3 & 0.574 & 0.568 &   -0.006 & 5861.5 &    * &  subsequently &       53.1 &      309.0 &  97.8 \\
      {*} &      primarily &       40.0 &       98.7 &  2602.8 & 0.586 & 0.573 &   -0.014 & 6504.5 &  *** &     primarily &       57.1 &      274.0 &  75.4 \\
      {*} &           fine &      353.2 &      488.9 &  2236.6 & 0.660 & 0.655 &   -0.005 & 5173.0 &   ns &        solely &       38.8 &      224.6 &  70.8 \\
      {*} &     accurately &       41.5 &       94.8 &  2179.0 & 0.618 & 0.608 &   -0.011 & 6059.5 &   ** &  particularly &      104.1 &      329.6 &  60.2 \\
      {*} &   specifically &      275.5 &      390.1 &  2018.9 & 0.557 & 0.564 &    0.007 & 4245.0 &   ns &   furthermore &      159.2 &      418.2 &  58.8 \\
      {*} &         solely &       30.4 &       73.9 &  1897.8 & 0.562 & 0.563 &    0.001 & 4791.0 &   ns &    accurately &       49.0 &      204.0 &  49.8 \\
\midrule
      {*} &           very &      334.8 &      233.5 &  1853.2 & 0.628 & 0.623 &   -0.005 & 5329.0 &   ns &          thus &      487.7 &       98.9 & 136.9 \\
      {*} &        jointly &       91.5 &       46.5 &  1527.9 & 0.604 & 0.617 &    0.014 & 3152.5 &  *** &         first &      540.8 &      177.2 &  94.0 \\
      {*} &           also &     1752.6 &     1542.8 &  1362.0 & 0.573 & 0.581 &    0.008 & 3816.0 &   ** &       usually &      236.7 &       33.0 &  84.5 \\
      {*} &        usually &      161.1 &      106.7 &  1137.4 & 0.618 & 0.622 &    0.004 & 4429.0 &   ns &          also &     1540.7 &      898.3 &  83.5 \\
      {*} &           well &      981.1 &      845.2 &  1031.8 & 0.535 & 0.525 &   -0.010 & 6211.0 &   ** &          much &      165.3 &       22.7 &  59.5 \\
      {*} &           thus &      451.1 &      364.2 &   945.7 & 0.624 & 0.620 &   -0.005 & 5090.5 &   ns &          very &      230.6 &       57.7 &  54.1 \\
      {*} &           much &      206.9 &      154.5 &   779.0 & 0.630 & 0.626 &   -0.003 & 5330.0 &   ns &       finally &      257.1 &       74.2 &  52.1 \\
       &  automatically &      173.8 &      127.9 &   713.6 & 0.626 & 0.618 &   -0.008 & 6221.5 &   ** &           far &      487.7 &      236.9 &  43.2 \\
      {*} &           then &      942.1 &      834.0 &   671.6 & 0.586 & 0.591 &    0.005 & 4287.5 &   ns &          only &     1106.0 &      756.1 &  32.2 \\
      {*} &       recently &      152.1 &      112.6 &   605.2 & 0.555 & 0.554 &   -0.001 & 5316.5 &   ns &        mainly &      151.0 &       45.3 &  29.2 \\
\bottomrule
\end{tabular}}
    \caption{Strongest changes in \textbf{adverb} usage.}
    \label{tab:app-adverbs}
\end{table*}

\begin{table*}
    \centering
    \scriptsize
    \begingroup
    \setlength{\tabcolsep}{5pt}
\begin{tabular}{rlrrr|lrrr}
\toprule
 % &                                                   lemma &  freq1 &  freq2 &     ll &                                             lemma &  freq1 &  freq2 &   ll \\
    \multicolumn{5}{c|}{\textbf{Naturalistic corpus}} & \multicolumn{4}{c}{\textbf{LLM-modified corpus}} \\
    \midrule
    & Target & ${t_1}$ & ${t_2}$ & \multicolumn{1}{c|}{$LL$} & Target & ${t_1}$ & ${t_2}$ & \multicolumn{1}{c}{$LL$} \\
\midrule
       &                        consistent with their intend use &         0.1 &        18.5 & 2440.5 &                         be important to note that &         8.2 &        72.1 & 28.6 \\
       &                         be consistent with their intend &         0.1 &        18.3 & 2410.8 &                           it be important to note &        10.2 &        72.1 & 25.6 \\
       &                 identify individual people or offensive &         0.1 &        17.8 & 2369.2 &                        important to note that the &         2.0 &        24.7 & 11.1 \\
       &                       information that name or uniquely &         0.1 &        17.7 & 2365.9 &                             it be worth note that &         8.2 &        39.1 & 10.8 \\
       &                          that name or uniquely identify &         0.1 &        17.7 & 2365.9 &                           can be find in appendix &         2.0 &        20.6 &  8.6 \\
       &                  or uniquely identify individual people &         0.1 &        17.8 & 2364.7 &                      paper be structure as follow &        10.2 &        37.1 &  7.9 \\
       &                  uniquely identify individual people or &         0.1 &        17.8 & 2364.7 &                     to enhance the performance of &         6.1 &        28.8 &  7.8 \\
       &                    name or uniquely identify individual &         0.1 &        17.8 & 2362.0 &                     the field of natural language &         2.0 &        18.5 &  7.4 \\
       &                  individual people or offensive content &         0.1 &        17.9 & 2357.6 &                        lead to the development of &         2.0 &        16.5 &  6.3 \\
       &                               the datum that be collect &         0.1 &        17.4 & 2325.2 &                             in the context of the &        10.2 &        33.0 &  6.2 \\
       &                             that be compatible with the &         0.4 &        17.5 & 2155.3 &              field of natural language processing &         2.0 &        14.4 &  5.1 \\
       &                          report the number of parameter &         0.2 &        15.8 & 2060.9 &                      the key contribution of this &         2.0 &        14.4 &  5.1 \\
       &                               of parameter in the model &         0.3 &        16.0 & 1954.4 &                           to address the issue of &         4.1 &        18.5 &  4.9 \\
       &                                the number of example in &         0.9 &        18.1 & 1945.2 &                        can be summarize as follow &        14.3 &        35.0 &  4.4 \\
       &                                 the source of the datum &         0.2 &        14.6 & 1847.9 &                         the paper be structure as &         8.2 &        24.7 &  4.3 \\
       &                                    be the source of the &         0.4 &        14.5 & 1738.0 &                           the extent to which the &         2.0 &        12.4 &  4.0 \\
       &                              number of parameter in the &         1.0 &        16.7 & 1702.7 &          demonstrate that our model significantly &         2.0 &        12.4 &  4.0 \\
       &                              the number of parameter in &         2.0 &        17.5 & 1435.6 &                        can be categorize into two &         2.0 &        12.4 &  4.0 \\
   {*} &                                 can be find in appendix &        14.1 &        34.6 &  906.0 &                        provide an overview of the &         4.1 &        16.5 &  3.9 \\
   {*} &                               be important to note that &         6.2 &        20.7 &  837.4 &                                in the case of the &         4.1 &        16.5 &  3.9 \\
   {*} &                                 it be important to note &         6.8 &        21.8 &  837.3 &                           it have be observe that &         4.1 &        14.4 &  3.0 \\
       &                      capability of large language model &         0.1 &         4.4 &  530.5 &           that our model significantly outperform &         4.1 &        14.4 &  3.0 \\
       &                          it be important to acknowledge &         0.2 &         3.5 &  373.8 &                              can be find in table &         4.1 &        14.4 &  3.0 \\
       &                       provide valuable insight into the &         0.1 &         3.0 &  318.8 &                 result demonstrate that our model &         2.0 &        10.3 &  2.9 \\
   {*} &                                   it be worth note that &        13.6 &        24.3 &  310.5 &                        this paper be structure as &         2.0 &        10.3 &  2.9 \\
   {*} &                           to enhance the performance of &         1.3 &         5.0 &  225.9 &                          to assess the quality of &         2.0 &        10.3 &  2.9 \\
   {*} &                                 can be attribute to the &         6.0 &        12.3 &  218.8 &          experimental result demonstrate that our &         2.0 &        10.3 &  2.9 \\
       &                        be important to acknowledge that &         0.1 &         2.0 &  215.7 &                           in the field of natural &         2.0 &        10.3 &  2.9 \\
   {*} &                          to assess the effectiveness of &         0.7 &         3.6 &  208.9 &                     between the source and target &         2.0 &        10.3 &  2.9 \\
   {*} &                              important to note that the &         1.8 &         5.6 &  203.7 &                         to mitigate the impact of &         2.0 &        10.3 &  2.9 \\
       &                                 grant fund by the korea &         1.0 &         4.1 &  202.0 &                    to assess the effectiveness of &         2.0 &        10.3 &  2.9 \\
       &                          particularly in the context of &         0.1 &         2.1 &  201.2 &                           in order to address the &         2.0 &        10.3 &  2.9 \\
       &                     natural science foundation of china &        11.9 &        19.8 &  199.7 &                     have show promising result in &         2.0 &        10.3 &  2.9 \\
       &                         ability of large language model &         0.1 &         2.0 &  194.8 &                    enhance the performance of the &         2.0 &        10.3 &  2.9 \\
       &                                  it be evident that the &         1.0 &         3.8 &  176.5 &                           can be attribute to the &         4.1 &        12.4 &  2.1 \\
       &                            prompt the model to generate &         0.1 &         1.8 &  174.0 &                  contribution can be summarize as &        12.2 &        24.7 &  2.1 \\
       &                  national natural science foundation of &        11.0 &        17.9 &  171.3 &                            be worth note that the &         2.0 &         8.2 &  2.0 \\
       &                                 may be attribute to the &         1.0 &         3.7 &  170.9 &                           be commonly refer to as &         2.0 &         8.2 &  2.0 \\
       &                                  we use the same prompt &         0.1 &         1.7 &  168.8 &                            model be train use the &         2.0 &         8.2 &  2.0 \\
       &                                   do not align with the &         0.3 &         2.2 &  152.9 &               result demonstrate that our propose &         2.0 &         8.2 &  2.0 \\
       \midrule
       &                                    the state of the art &        11.4 &         3.9 &  390.6 &                          the paper be organize as &        22.4 &         2.1 &  9.7 \\
   {*} &                             paper be organize as follow &        11.0 &         4.8 &  255.2 &                       paper be organize as follow &        30.6 &         6.2 &  8.6 \\
       &                         acknowledgment we would like to &         6.9 &         2.6 &  205.0 &                                have be show to be &        20.4 &         2.1 &  8.5 \\
       &                             the encoder and the decoder &         4.8 &         1.4 &  193.5 &                              it have be show that &        22.4 &         4.1 &  6.8 \\
       &                                  we would like to thank &        22.6 &        14.5 &  180.1 &                         the good of our knowledge &       114.3 &        68.0 &  5.8 \\
       &                          and the anonymous reviewer for &         5.2 &         1.8 &  179.2 &                                to the good of our &       114.3 &        70.0 &  5.2 \\
   {*} &                              result show that our model &         4.4 &         1.4 &  163.4 &                          of the paper be organize &        14.3 &         2.1 &  5.0 \\
       &                        the anonymous reviewer for their &        24.4 &        16.4 &  161.1 &                            this work be as follow &        14.3 &         2.1 &  5.0 \\
       &                                in this paper we present &         4.1 &         1.3 &  155.9 &                      contribution of this work be &        24.5 &         8.2 &  4.1 \\
   {*} &                       experimental result show that our &         9.2 &         4.6 &  155.7 &                       contribution of our work be &        12.2 &         2.1 &  3.9 \\
   {*} &                                the paper be organize as &         6.4 &         2.8 &  155.5 &                                of this work be as &        12.2 &         2.1 &  3.9 \\
       &                                    our model be able to &         4.3 &         1.4 &  152.8 &                    the contribution of this paper &        12.2 &         2.1 &  3.9 \\
       &                                we evaluate our model on &         6.6 &         2.9 &  150.0 &               contribution be summarize as follow &        16.3 &         4.1 &  3.8 \\
       & \scalebox{.82}[1.0]{bidirectional encoder representations from transformers} &         5.7 &         2.3 &  149.1 &                     contribution of this paper be &        24.5 &        10.3 &  2.9 \\
       &                               the source and the target &         3.9 &         1.2 &  147.3 &                        result show that our model &        14.3 &         4.1 &  2.9 \\
       &                                in this paper we propose &         2.8 &         0.7 &  137.6 &                            from the point of view &        10.2 &         2.1 &  2.9 \\
       &                                    rest of the paper be &         5.6 &         2.4 &  133.8 &                              the point of view of &        10.2 &         2.1 &  2.9 \\
       &                               we use the adam optimizer &         6.0 &         2.7 &  130.5 &                              which be base on the &        10.2 &         2.1 &  2.9 \\
       &                                     state of the art in &         4.2 &         1.6 &  126.4 &                               it be not clear how &        10.2 &         2.1 &  2.9 \\
       &                             language model such as bert &         5.2 &         2.2 &  125.1 &                                that can be use to &        16.3 &         6.2 &  2.3 \\
       &                                   the rest of the paper &         8.0 &         4.1 &  124.8 &                     main contribution of our work &        12.2 &         4.1 &  2.1 \\
       &                                 state of the art result &         1.8 &         0.3 &  124.3 &                                be one of the most &        12.2 &         4.1 &  2.1 \\
       &                                     we can see that the &        15.8 &        10.3 &  119.8 &                      we conduct experiment on the &        12.2 &         4.1 &  2.1 \\
       &                                 and future work in this &         6.8 &         3.4 &  119.1 &                         on the performance of the &        12.2 &         4.1 &  2.1 \\
   {*} &                                of the paper be organize &         4.6 &         1.9 &  118.4 &                     contribution in this paper be &         8.2 &         2.1 &  1.9 \\
       &                                      of the word in the &         5.1 &         2.2 &  118.1 &                we make the following contribution &         8.2 &         2.1 &  1.9 \\
       &                                    new state of the art &         3.0 &         0.9 &  118.0 & \scalebox{.82}[1.0]{experimental result demonstrate the effectiveness} &         8.2 &         2.1 &  1.9 \\
       &                                     on the basis of the &         5.6 &         2.7 &  112.0 &                         an answer to the question &         8.2 &         2.1 &  1.9 \\
       &                               future work in this paper &         5.2 &         2.4 &  108.0 &                           for the purpose of this &         8.2 &         2.1 &  1.9 \\
       &                      reviewer for their helpful comment &         6.1 &         3.1 &  104.2 &                         this be the first attempt &         8.2 &         2.1 &  1.9 \\
       &                        thank the anonymous reviewer for &        17.7 &        12.2 &  101.9 &                           be the first attempt to &         8.2 &         2.1 &  1.9 \\
       &                                     state of the art on &         3.1 &         1.1 &   99.5 &                                we be the first to &        34.7 &        20.6 &  1.8 \\
       &                                 the hidden state of the &         6.5 &         3.4 &   98.5 &                      the main contribution of our &        14.3 &         6.2 &  1.6 \\
   {*} &                               this paper be organize as &         4.3 &         1.9 &   98.0 &                 experimental result show that our &        10.2 &         4.1 &  1.3 \\
       &                               we compare our model with &         5.6 &         2.8 &   96.0 &                         to encourage the model to &        10.2 &         4.1 &  1.3 \\
       &                          show that our model outperform &         2.5 &         0.8 &   96.0 &                            similar to the one use &        10.2 &         4.1 &  1.3 \\
       &                    anonymous reviewer for their helpful &         7.6 &         4.3 &   95.6 &                   evaluate the performance of the &        10.2 &         4.1 &  1.3 \\
       &                                 would like to thank the &        13.3 &         8.8 &   94.3 &                    main contribution of this work &        10.2 &         4.1 &  1.3 \\
       &                   acknowledgment we thank the anonymous &         2.7 &         0.9 &   92.5 &                     the main contribution of this &        24.5 &        14.4 &  1.3 \\
       &                                    for each word in the &         3.4 &         1.4 &   87.8 &             effectiveness of the propose approach &         6.1 &         2.1 &  1.0 \\
\bottomrule
\end{tabular}
    \endgroup
    \caption{Strongest changes in the use of word 5-grams. Shown columns: \tone: frequency in \tone per million words; \ttwo: frequency in \ttwo per million words; $LL$: log-likelihood score.} %: top 40 increases (top) and decreases (bottom) based on log-likelihood values. \tone: frequency in \tone per million words; \ttwo: frequency in \ttwo per million words; $LL$: log-likelihood score. Naturalistic corpus targets marked with an asterisk also appear in the top 400 strongest changes in the LLM-modified corpus (for the corresponding increase/decrease direction).}
    \label{tab:app-fivegrams}
\end{table*}

\clearpage
\begin{table*}
    \centering
    % \resizebox{\linewidth}{!}{%
    % \begin{tabular}{rcl}
    \scriptsize
    \begin{tabular}{>{\raggedleft\arraybackslash}p{0.43\linewidth}c>{\raggedright\arraybackslash}p{0.43\linewidth}}
    \toprule
    % 431 sents -- 65% t1 / 35% t2 -- clust 7
    measure the semantic similarity between generated questions to & \textbf{ensure} & that the questions assess the same content . \\
    We add an adversarial objective to & \textbf{ensure} & that the model focuses on language - agnostic \\
    sampling process , we store this data to & \textbf{ensure} & that every model processes exactly the same set \\
    we diversify the output of each task to & \textbf{ensure} & that the model can provide a variety of \\
    \midrule
    % 363 sents -- 43% t1 / 57% t2 -- clust 4
    To & \textbf{ensure} & the high quality of the annotation procedure , \\
    To & \textbf{ensure} & the quality of data , they also calculated \\
    Finally , to & \textbf{ensure} & training stability and prevent overfitting , we modify \\
    the hyperparameters the same for different models to & \textbf{ensure} & the fairness of the experiment . \\
    \bottomrule
    \end{tabular}%}
    % \caption{Top: 65\%\,\tone vs.\ 35\%\,\ttwo (431 sentences). Bottom: 43\%\,\tone vs.\ 57\%\,\ttwo (363 sentences).}
    \caption{Example uses of the verb \textit{ensure}. 
    Top cluster: broad sense of finality similar to conjunctions like \textit{in order to} (65\%\,\tone vs.\ 35\%\,\ttwo; 431 sentences). 
    Bottom cluster: more specific sense `to guarantee'
    (43\%\,\tone vs.\ 57\%\,\ttwo; 363 sentences).
    The growing use of the more specialized second sense is consistent with an increasing neighborhood density ($\Delta ND=0.013$).}
    \label{tab:clust-ensure}
\end{table*}

\begin{table*}
    \centering
    % \resizebox{\linewidth}{!}{%
    % \begin{tabular}{rcl}
    \scriptsize
    \begin{tabular}{>{\raggedleft\arraybackslash}p{0.43\linewidth}c>{\raggedright\arraybackslash}p{0.43\linewidth}}
    \toprule
    % 281 sents -- 60% t1 / 40% t2 -- clust 6
    The latter aims to & \textbf{utilize} & the multi - level interests to enhance both \\
    of data augmentation method and proposed methods to & \textbf{utilize} & the augmented data in more effective ways ( \\
    introduced pivot - based transfer learning techniques to & \textbf{utilize} & the resources of the pivot language . \\
    We first transform emoticons into textual information to & \textbf{utilize} & their rich emotional information . \\
    \midrule
    % 354 sents -- 44% t1 / 56% t2 -- clust 0
    Finally , we & \textbf{utilize} & a ridge regression classifier to obtain final classification \\
    For both tasks , we & \textbf{utilize} & a publicly available implementation that has been trained \\
    In the second setting , we & \textbf{utilize} & the MLM training objective inherited from PLMs to \\
    % Finally , we & \textbf{utilize} & these generated semantic fragments to reconstruct the semantic \\
    Subsequently , we & \textbf{utilize} & a language model equipped with adapters to obtain \\
    \bottomrule
    \end{tabular}%}
    % \caption{Top: 60\%\,\tone vs.\ 40\%\,\ttwo (281 sentences). Bottom: 44\%\,\tone vs.\ 56\%\,\ttwo (354 sentences).}
    \caption{Example uses of the verb \textit{utilize}.
    Top cluster: more specific sense `use to the fullest potential' 
    (60\%\,\tone vs.\ 40\%\,\ttwo; 281 sentences).
    Bottom cluster: more general sense `make use of' 
    (44\%\,\tone vs.\ 56\%\,\ttwo; 354 sentences).
    The growing use of the broader second sense is consistent with a falling neighborhood density ($\Delta ND=-0.026).$}
    \label{tab:clust-utilize}
\end{table*}

\begin{table*}
    \centering
    % \resizebox{\linewidth}{!}{%
    % \begin{tabular}{rcl}
    \scriptsize
    \begin{tabular}{>{\raggedleft\arraybackslash}p{0.43\linewidth}c>{\raggedright\arraybackslash}p{0.43\linewidth}}
    \toprule
    % 146 sents -- 58% t1 / 42% t2 -- clust 2
    , word order difference is one of the & \textbf{crucial} & factors that impact cross - lingual transfer ( \\
    Thus , sarcasm detection might be the first & \textbf{crucial} & step in these systems . \\
    Apart from these & \textbf{crucial} & extensions we made for providing the use of \\
    of styles used makes measurements , a most & \textbf{crucial} & aspect of scientific writing , challenging to extract \\
    \midrule
    % 208 sents -- 38% t1 / 62% t2 -- clust 0
    Therefore , it is & \textbf{crucial} & to understand the speaker 's intentions and emotions \\
    of the inherent individual differences , it is & \textbf{crucial} & to incorporate diversity in the data collection process \\
    % and analysis of the reviews , it is & \textbf{crucial} & to properly respond to the user . \\
    the annotation tasks automatically , we deem it & \textbf{crucial} & to expand our proposed framework to new scenarios \\
    These challenges make it & \textbf{crucial} & to devise a new framework that can work \\
    \bottomrule
    \end{tabular}%}
    % \caption{Top: 58\%\,\tone vs.\ 42\%\,\ttwo (146 sentences). Bottom: 38\%\,\tone vs.\ 62\%\,\ttwo (208 sentences).}
    \caption{Example uses of the adjective \textit{crucial}.
    Top cluster: finality-connoted sense `important in determining an outcome' 
    (58\%\,\tone vs.\ 42\%\,\ttwo; 146 sentences). 
    Bottom cluster: more general sense `important, significant' 
    (38\%\,\tone vs.\ 62\%\,\ttwo; 208 sentences).
    The growing use of the broader second sense is consistent with a falling neighborhood density ($\Delta ND=-0.018$).}
    \label{tab:clust-crucial}
\end{table*}

\begin{table*}
    \centering
    %\resizebox{\linewidth}{!}{%
    %\begin{tabular}{rcl}
    \scriptsize
    \begin{tabular}{>{\raggedleft\arraybackslash}p{0.43\linewidth}c>{\raggedright\arraybackslash}p{0.43\linewidth}}
    \toprule
    % 171 sents -- 63% t1 / 37% t2 -- clust 6
    a mixture of content and structural properties , & \textbf{notably} & the systems in the 2016 document alignment shared \\
    , the model shows diminished retrieval performance , & \textbf{notably} & in terms of R@1 . \\
    had been dominated by the statistical methods , & \textbf{notably} & the phrase - based ( Koehn et al \\
    switching corpora in comparison with monolingual corpora , & \textbf{notably} & for high - resource languages , e.g. , \\
    \midrule
    % 309 sents -- 44% t1 / 56% t2 -- clust 1
    predictions for the labels of other pairs is & \textbf{notably} & lower . \\
    % that their performance on our test dataset is & \textbf{notably} & subpar ( < 10 \% ) . \\
    and GPT - labels , however , is & \textbf{notably} & larger than the other models , with a \\
    and XLM - RoBERTa - Large models are & \textbf{notably} & improved in both the full and partial benchmarks \\
    DSP on most datasets , where it is & \textbf{notably} & more efficient to train . \\
    \bottomrule
    \end{tabular}%}
    % \caption{Top: 63\%\,\tone vs.\ 37\%\,\ttwo (171 sentences). Bottom: 44\%\,\tone vs.\ 56\%\,\ttwo (309 sentences).}
    \caption{Example uses of the adverb \textit{notably}.
    Top cluster: semantically broad usage typically introducing an example, similar to `especially, particularly'
    (63\%\,\tone vs.\ 37\%\,\ttwo; 171 sentences). 
    Bottom cluster: intensifier-like sense `very'
    (44\%\,\tone vs.\ 56\%\,\ttwo; 309 sentences).
    The intensification use is restricted to a relatively small set of cooccurrents (typically gradable adjectives), which aligns with an increasing neighborhood density ($\Delta ND=0.029$).}
    \label{tab:clust-notably}
\end{table*}

\begin{table*}[t]
\centering
\begin{tabular}{ll}
\toprule
\textbf{Dimension} & \textbf{Question} \\
\midrule
\multirow{2}{*}{Clarity} 
& I read this text smoothly and fluently. \\
& I was able to understand the paragraph without difficulty. \\

\midrule
\multirow{2}{*}{Authenticity}
& I felt a sense of connection with the researchers through their writing. \\
& The authors are genuinely engaged with the topic they studied. \\

\midrule
\multirow{2}{*}{Trustworthiness}
& The authors appeared competent and trustworthy. \\
& I remained critical of the authors’ arguments while reading. \\

\midrule
\multirow{2}{*}{Excitement}
& I am excited to read this paper in more depth. \\
& I found the text enjoyable to read. \\
\bottomrule
\end{tabular}
\caption{The following questions are used to measure four different dimensions of reading experience.}
   \label{tab:readingexperience}
\end{table*}

\end{document}